\def\arxivfmt{1} %
\def\fullpage{1} %

\if\arxivfmt 1 
\documentclass[11pt,x11names]{article}
\usepackage{setspace}
\else
\fi

\if\arxivfmt 0
\usepackage[nonatbib]{neurips_2021}
\fi

\if\arxivfmt 2
\usepackage{aistats2022}
\fi 

\if\arxivfmt 1
\usepackage{fullpage}
\usepackage{authblk}
\fi
 
\usepackage{amsmath, latexsym}
\usepackage{amscd}
\usepackage{amsbsy}
\usepackage{amssymb}
\usepackage{amsthm}
\usepackage{todonotes}
\usepackage{comment} 
\usepackage{caption} 
\usepackage{subcaption} 

\usepackage{nicefrac}

\newtheorem{theorem}{Theorem}
\newtheorem*{theorem*}{Theorem}
\newtheorem*{corollary*}{Corollary}

\newtheorem{innercustomthm}{Theorem}
\newenvironment{theoremlit}[1]
{\renewcommand\theinnercustomthm{#1}\innercustomthm}
{\endinnercustomthm}
\usepackage{xcolor}
\usepackage{hyperref}
\hypersetup{
	colorlinks,
	linkcolor={red!60!black},
	citecolor={blue!60!black},
	urlcolor={blue!80!black}
}
\usepackage{cleveref}
\crefname{theoremlitCounter}{Theorem}{Theorems}

\crefname{assumption}{assumption}{assumptions}

\newtheorem{lemma}{Lemma}
\newtheorem{corollary}[theorem]{Corollary}
\theoremstyle{definition}
\newtheorem{assumption}{Assumption}
\newtheorem{definition}{Definition}
\newtheorem{remark}{Remark}
\newtheorem{example}{Example}
\newtheorem{proposition}{Proposition}
\usepackage{booktabs}
\usepackage{url}
\usepackage[numbers]{natbib}
\usepackage{outline}

\usepackage{enumitem}

\usepackage[Algorithm,ruled]{algorithm}
\usepackage[noend]{algpseudocode} 

\usepackage{tikz}
\usetikzlibrary{positioning,arrows}
\newcommand{\pr}{\mathbb{P}}
\newcommand{\norm}[1]{\left\lVert#1\right\rVert}

\usepackage{shortcuts}
\usepackage{amssymb}%
\usepackage{pifont}%
\usepackage[symbol]{footmisc}

\newcommand{\vc}{d_{\op{vc}}}

\newcommand{\IPW}{\op{IPW}}
\newcommand{\DM}{\op{DM}}

\newcommand{\MIS}{\op{MIS-mw}}
\newcommand{\FQE}{\op{FQE}}
\newcommand{\DRL}{\op{DRL-mw}}

\newcommand{\DR}{\op{DR}}

\newcommand{\Rmax}{R^{\max}}
\newcommand{\nS}{\vert \mathcal S\vert}
\newcommand{\Ss}{\mathcal S}
\newcommand{\Aa}{\mathcal A}
\newcommand{\Xx}{\mathcal X}
\newcommand{\margP}{\tilde P}
\newcommand{\margR}{\tilde R}
\newcommand{\hatmargR}{\hat{\tilde{R}}}
\newcommand{\Yy}{\mathcal Y}
\newcommand{\hatmargP}{\widehat{{P}}}
\newcommand{\tpi}{\pi}

\newcommand{\outcomeratio}{\Delta\mu}

\newcommand{\pyax}{P(Y=y\mid x,a)}
\newcommand{\pyaX}{P(Y=y\mid a,X)}

\newcommand{\pypi}{P(Y=y\mid\pi)}
\newcommand{\hpypi}{\hat P(Y=y\mid\pi)}
\newcommand{\py}[1]{P(Y=#1\mid \pi)}

\newcommand{\hpi}[1]{\hat\pi^*_{#1}}
\newcommand{\spi}[1]{\pi^*_{#1}}
\newcommand{\etaax}[1]{\mu(1\mid#1, x)}

\newcommand{\Vpi}{V^\pi} 
\newcommand{\tildeV}{\tilde{V}} 
\newcommand{\hatV}{\widehat{V}} 
\newcommand{\htV}{\widehat{{V}}}
\newcommand{\htpiV}{\widehat{{V}}^{\tpi}}
\newcommand{\htPiV}[1]{\widehat{{V}}^{#1}}
\newcommand{\htPiQ}[1]{\widehat{{Q}}^{#1}}
\newcommand{\tildeQ}{\widetilde{Q}} 
\newcommand{\htQ}{\widehat{{Q }}}
\newcommand{\tildeR}{\tilde{R}} 
\newcommand{\horizon}{T}

\usepackage{tikz}
\newcommand*\circled[1]{\tikz[baseline=(char.base)]{
		\node[shape=circle,draw,inner sep=2pt] (char) {#1};}}
\newcommand{\1}{\circled{1}}
\newcommand{\2}{\circled{2}}

\usepackage{empheq}
\usepackage{xcolor}
\makeatletter
\newcommand*{\defeq}{\mathrel{\rlap{%
			\raisebox{0.3ex}{$\m@th\cdot$}}%
		\raisebox{-0.3ex}{$\m@th\cdot$}}%
	=}
\makeatother

\usepackage{footnote}
\usepackage{footmisc}

\renewcommand{\thefootnote}{\alph{footnote}}

\newcommand{\astfootnote}[1]{%
	\let\oldthefootnote=\thefootnote%
	\setcounter{footnote}{0}%
	\renewcommand{\thefootnote}{\fnsymbol{footnote}}%
	\footnote{#1}%
	\let\thefootnote=\oldthefootnote%
}

\if\arxivfmt 1
\title{Stateful Offline Contextual Policy Evaluation and Learning}
\author{
	Nathan Kallus and Angela Zhou\footnote{Corresponding author. \url{angela-zhou@berkeley.edu}. Authors listed in alphabetical order.}
}
\fi

\if\arxivfmt 0
\title{Stateful Offline Contextual Policy Evaluation and Learning}
\fi

\begin{document}
	
\if\arxivfmt 1
	\maketitle

	\fi
	
	\if\arxivfmt 0
	\maketitle
	\fi
	
\if\arxivfmt 2
\twocolumn[

\aistatstitle{Stateful Offline Contextual Policy Evaluation and Learning}
]
\fi 
	
	\begin{abstract}
		We study off-policy evaluation and learning from sequential data in a structured class of Markov decision processes that arise from repeated interactions with an exogenous sequence of arrivals with contexts, which generate unknown individual-level responses to agent actions. This model can be thought of as an offline generalization of contextual bandits with resource constraints. We formalize the relevant causal structure of problems such as dynamic personalized pricing and other operations management problems in the presence of potentially high-dimensional user types. The key insight is that an individual-level response is often not causally affected by the state variable and can therefore easily be generalized across timesteps and states. When this is true, we study implications for (doubly robust) off-policy evaluation and learning by instead leveraging single time-step evaluation, estimating the expectation over a single arrival via data from a population, for fitted-value iteration in a \textit{marginal MDP}. We study sample complexity and analyze error amplification that leads to the persistence, rather than attenuation, of confounding error over time. In simulations of dynamic and capacitated pricing, we show improved out-of-sample policy performance in this class of relevant problems.
	\end{abstract}
	
		\section{Introduction}
Offline reinforcement learning seeks to reuse existing data to evaluate and learn novel policies and is crucial in applications with limited freedom to experiment but plentiful logged data. In general Markov decision processes (MDPs), offline reinforcement learning can be very difficult, as we must understand the effect of actions in each state and time, whether in model-based (e.g., learn the transition kernel) or model-free methods (e.g., learn $Q$-functions). However, many practically-relevant problems fit in simpler, more tractable classes of MDPs with ``sequential decision-making" but not ``longitudinal data", for example because transitions arise in a stochastic system from exogenous arrivals. In this paper, we study off-policy evaluation and optimization from observational data in this special class. At each timestep the \emph{same} contextual response model generates both transitions and rewards. The setting is
a variant of offline contextual bandits with constraints, where 
the same randomness generates transitions in the system state (status of the constraints) and rewards in the system. 
We call this setting, common in operations management,  ``stateful"
to emphasize the well-understood and simple system state, like inventory state, in contrast to the unknown potentially high-dimensional 
``contextual"
response model, like an individual's propensity to purchase, that must be learned.

We first describe some stylized examples to illustrate how previously studied classical problems in fact share this broader structure: consider personalized dynamic pricing with inventory constraints, or managing a rideshare system and repositioning vehicles by making price offers to individuals. The {system state} includes capacities of each resource, or locations of cars in the system. Individuals with contexts (covariates) arrive exogenously. The system takes actions, such as personalized price or trip offers. Given a context and action, the individual response changes system dynamics: the purchase of a product consumes resources, or accepting a price offer and ride from one location to another moves cars. 
But given that we can offer the resource at all, the state of the system does not further affect the response except by affecting our pricing decisions.

We focus on the evaluation and optimization of state-dependent policies from offline trajectories collected from these system dynamics. Confounding is of particular concern in such observational data. Naturally, system actions can be spuriously correlated with outcomes. For example, we expect observational operational data to bias toward higher-revenue actions: higher price offers are made to individuals deemed more likely to accept. However, in our setting, the underlying system state does not causally affect the individual response and therefore, unlike individual-level covariates, the system state is not a confounder, making it easier to learn the response model from observational data, and then use it to design sequential policies. Ultimately we show how this special structure of problems, common to operations problems arising from uniformized stochastic systems permits developing specialized OPE.

The contributions of this paper are as follows. We model the above structure and 
study 
estimation of the transition probabilities in a lifted \textit{marginal} MDP via single timestep off-policy evaluation. Therefore we reduce the analysis of a high-dimensional, continuous state space MDP to the standard tabular analysis of our marginal MDP, reducing the number of nuisances from $T+1$ for doubly-robust OPE to \textit{just two} in our setting. We show $\widetilde O\left( \nicefrac{  \horizon^3 }{\epsilon^2} \right)$ trajectories are required for off-policy optimization to achieve $\epsilon$-suboptimal value, where $\horizon$ is the horizon (omitting logarithmic terms).
We study error amplification for the dynamic and capacitated pricing example and show that bias from naive model-based approaches would generally persist in realistic scenarios.
We validate the theory and structural analysis in simulations
where we improve on naive model-based approaches and generic offline RL.

	\section{Problem setup: Stateful off-policy evaluation and learning}\label{problem-setup}
We first describe the generic full-information MDP that generates our data before describing the restrictions that characterize the stateful setting. For ease of reference, we partition the state space of the full-information MDP into a product space of the \textit{discrete system state space} $\Ss$, potentially continuous \textit{context/covariate space} $\Xx$, 
and discrete covariate-conditional response space $\Yy$:
$\;\Ss \times \Xx \times \Yy$. The inclusion of $Y$ in the state variable is purely informational.
Consider a finite-horizon setting with $T+1$ timesteps, and denote the initial system state $s_0$; timesteps are indexed $0, \dots,T$. Uppercase ($S,X$) indicates random variables; lowercase ($s,x$) fixed values; and prime ($s',x'$) next-timestep values. Let $\Aa(s,x)$ denote the discrete action space feasible from the state $(s,x)$.
 A contextual policy $\pi_t \colon \Ss \times \Xx \mapsto \Delta^{\Aa}$ maps from system state/context to a distribution over actions, where $\Delta^{\Aa}$ is the set of distributions defined on $\Aa$, so that $\pi_t(a\mid s,x) =
 { \pr(A=a \mid S=s, X=x)}
 	$ gives the probability of taking action $a$ given state and context information. 
Let $\tpi = \{ \pi_t\}_{t=0,\dots,T}
$
denote the MDP policy
in a function class $ \Pi_{0:T}$. 
 	Reward is a {known} deterministic function of next state transition, $R(s,a,s')$.

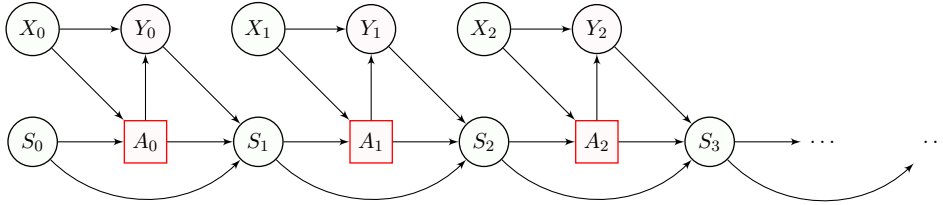
\begin{figure*}
	\centering
	\begin{tikzpicture}[%
		>=latex',node distance=2cm, minimum height=0.75cm, minimum width=0.75cm,
		state/.style={draw, shape=circle, draw=black, fill=green!2, line width=0.5pt},
		context/.style={draw, shape=circle, draw=black, fill=purple!2, line width=0.5pt},
		action/.style={draw, shape=rectangle, draw=red, fill=red!2, line width=0.5pt},
		reward/.style={draw, shape=rectangle, draw=blue, fill=blue!2, line width=0.5pt},
		scale = 0.75, transform shape
		]
		\node[state] (S0) at (0,0) {$S_0$};
		\node[action,right of=S0] (A0) {$A_0$};
		\node[state,above of =S0] (x0) {$X_0$};
		\node[context,above of =A0] (y0) {$Y_0$};
		\node[state,right of=A0] (S1) {$S_1$};
		\node[action,right of=S1] (A1) {$A_1$};
		\node[state,above of =S1] (x1) {$X_1$};
		\node[context,above of =A1] (y1) {$Y_1$};
		\node[state,right of=A1] (S2) {$S_2$};
		\node[action,right of=S2] (A2) {$A_2$};
		\node[state,above of =S2] (x2) {$X_2$};
		\node[context,above of =A2] (y2) {$Y_2$};
		\node[state,right of=A2] (S3) {$S_3$};
		\node[right of=S3] (dots) {$\ldots$};
		\node[right of=dots] (dots2) {$\ldots$};
		\draw[->] (S0) -- (A0);
		\draw[->] (x0) -- (A0);
		\draw[->] (A0) -- (y0);
		\draw[->] (x0) -- (y0);
		\draw[->] (y0) -- (S1);
		\draw[->] (A0) -- (S1);
		\draw[->] (S0) edge[bend right=45] (S1);
		\draw[->] (S1) -- (A1);
		\draw[->] (x1) -- (A1);
		\draw[->] (A1) -- (y1);
		\draw[->] (x1) -- (y1);
		\draw[->] (y1) -- (S2);
		\draw[->] (A1) -- (S2);
		\draw[->] (S1) edge[bend right=45] (S2);
		\draw[->] (S2) -- (A2);
		\draw[->] (x2) -- (A2);
		\draw[->] (A2) -- (y2);
		\draw[->] (x2) -- (y2);
		\draw[->] (y2) -- (S3);
		\draw[->] (A2) -- (S3);
		\draw[->] (S2) edge[bend right=45] (S3);
		\draw[->] (S3) -- (dots);
		\draw[->] (S3) edge[bend right=45] (dots2);
	\end{tikzpicture}
	\caption{The ``stateful" decision model we consider; rewards are functions of $S_t,A_t,S_{t+1}$.}\label{diag-stateful-policy-learning}
\end{figure*}

\paragraph{``Stateful contextual" structure.}
We next specify the restrictions on this MDP that give rise to our ``stateful'' setting. These are illustrated in \Cref{diag-stateful-policy-learning}. Roughly: contexts arrive exogenously and contextual responses $Y$ come from a stationary conditional distribution ${\pr(Y\mid  X,A)}$ and deterministically generate the system transitions.
 We henceforth use the shorthand 
 ${P(s',x',y\mid s,x,y_{-1},a)}$ for the transition model under this convention, although dependence on $y_{-1}$ is purely artificial/notational and will be omitted in general.
 
We formalize as assumptions the general structure that appears commonly in more specialized problem contexts elsewhere that describes the ``stateful" setting.
\begin{assumption}[Exogenous context process]\label{asn-X-exogeneous-context} The transition factorizes as
	\if \fullpage 1
\[{P(s',x',y\mid s,x,y_{-1},a) = P(s',y\mid s,x,a) f(x') \;\; \forall s,s',x,x',y_{-1},y,a}\]
\fi 
\if \fullpage 0 
\begin{align*}{P(s',x',y\mid s,x,y_{-1},a) &= P(s',y\mid s,x,a) f(x') \;\; \\
		&\forall s,s',x,x',y_{-1},y,a}\end{align*} 
\fi 
\end{assumption}

	\begin{assumption}[Contextual-response transitions]\label{asn-lin-mdp-pot-out}

We know  $s'(s,y): \Ss\times\Yy \mapsto \Ss$ such that when $s'$ is not absorbing from $s$ we have:
		\begin{align*}	 
&P(s',y\mid s,x,a)=\delta_{s'-s'(s,y)} P(Y=y\mid x,a)
		\end{align*}
	\end{assumption}
We can easily extend to random transitions given responses, but focus on deterministic for concreteness and as it captures the most relevant application settings. \Cref{asn-X-exogeneous-context} 
arises from contextual bandits or uniformizing (with contexts) a stochastic system \cite{gallego2019revenue,meyn2008control}. \Cref{asn-lin-mdp-pot-out} reflects the offline contextual bandit nature of the problem and encodes that $Y_t$ is independent of the originating state.
\cite{el2020lookahead} leverages a factorization with exogenous information, but not a contextual response model and notes that the ``system transition function" construction is the norm in control/operations research \cite{bertsekas1996neuro,powell2007approximate}.

For ease of presentation we introduce $\{ s',y\mid s,a \} $ as the pairs of next states and contextual responses reachable from $s$.

\begin{definition}[Reachable state transition-potential outcomes]\label{def-reachablepotout}
	\if \fullpage 1 
$$\{ (s',y) \mid s\} \defeq \{ (s',y)  \in \Ss \times \Yy
\colon 
\exists 
\;
x,a \text{ s.t. } 
P(s',y \mid s,x,a) > 0
 \} $$
 \fi 
 	\if \fullpage 0
\begin{align*}&\{ (s',y) \mid s\}\defeq \\
	& \{ (s',y)  \in \Ss \times \Yy
 	\colon 
 	\exists 
 	\;
 	x,a \text{ s.t. } 
 	P(s',y \mid s,x,a) > 0
 	\} 
 	\end{align*} 
 	\fi 
	\end{definition}

	The corresponding full-information MDP is $\mathcal M = (\Ss \times \Xx \times \Yy, \Aa, P, R, \horizon)$ where $P$ is the full-information transition kernel. Our \textit{observational} data comprises of $N$ trajectories; denote individual observations as $S_t^{i}$ for $t$ timestep of trajectory $i$: 	$ 
	\textstyle
	\{ S_t^{i}, X_t^{i}, A_t^{i}, Y_t^{i},R_{t}^{i}
	\}_{0:T}^n$.

Without loss of generality we omit the $Y_{t-1}$ information state from the full-information \textit{value/reward to go} $V$- and state-action value $Q$-function, since ${V_t(s,x,y_{-1})=V_t(s,x)}$: 
\if \fullpage 1 
	 $$\textstyle
	 { \Vpi_{t}(s,x)=\mathbb{E}_{ \pi}\left[\sum_{t'=t}^{\horizon}
		R_{t'}
\mid  S_{t}=s, X_{t}=x\right]}, 
\;\;
Q^\pi_{t}(s,x,a) =  \E
	\bracks{ \sum_{t'=t}^{\horizon} R_{t'} \mid S_t = s, X_t =x,A_t =a }$$
	\fi 
	\if \fullpage 0 
\begin{align*} 
	 \Vpi_{t}(s,x)&\textstyle=\mathbb{E}_{ \pi}\left[\sum_{t'=t}^{\horizon}
		R_{t'}
		\mid  S_{t}=s, X_{t}=x\right], \\
	Q^\pi_{t}(s,x,a) &\textstyle=  \E
	\bracks{ \sum_{t'=t}^{\horizon} R_{t'} \mid S_t = s, X_t =x,A_t =a }
	\end{align*} 
	\fi

It is useful to define the \textit{context-marginalized} value function  $\tildeV^\pi_t (s) = \E[V^\pi(s, X) ]$, the value function at system state $S_t$ marginalized over context distribution $X_t$, and analogously $\tildeQ$. Under \cref{asn-X-exogeneous-context,asn-lin-mdp-pot-out} and the notation of defn. 1, the $Q$-function in the full-information MDP is:
\begin{equation} 
 Q_t^\pi(s,x,a) = \sum_{(s',y)\mid s} \pyax (R(s,a,s') + \tildeV_{t+1}^\pi(s'))
 \label{eqn-qfunction}
\end{equation} 
Finally, throughout the paper we will assume the behavior policy is \textit{stationary}, and \textit{not} history-adapted for a simpler statement of our results.\footnote{We leave finer-grained analysis of dependent data, where the benefits of pooling data also trade-off against dependence and mixing rates, for future work; \Cref{cor-nuisance-rates} highlights how this assumption may be removed via standard arguments.} 
\begin{assumption}[Stationary behavior policy]\label{asn-stationary-behavior-policy}
	${\pi^b_t(a\mid s,x)}$ is stationary: possibly time-varying, but not history-dependent upon $\{S_{t'},X_{t'},A_{t'},Y_{t'}\}_{t'<t}$.
\end{assumption}

\textbf{Specific examples of stateful problems.}\label{sec-model-based-approach} 
We discuss illustrative examples. The first
 example, 
 single-item personalized and dynamic pricing with inventory constraints,
 is adapted from classical models for network revenue management (\cite{gallego2019revenue,gallego1997multiproduct}).\footnote{Instead of assuming arrivals would deterministically purchase and setting the decision variable to be fare availability, we consider a stochastic demand response model.} 
\begin{example}[Single item personalized and dynamic pricing]\label{ex-singlenrm}

	$Y_t \in \{0,1\}$ is purchase/no-purchase, respectively, and $A_t \in \{0,1\}$ is whether a discount of value $d$ is not or is offered. Let $p(a)$ be the price corresponding to taking action $A=a$. The reward is fixed given transition to $s'$: $R(s,a,s') = p(a) \mathbb{I}[y  = 1] \mathbb{I}[s>0].$ For short let $R(a)$ denote price of product under $a$, e.g. reward only received if item is sold, and we can only sell if we have stock so $s'(s,y) = {\mathbb{I}[s>0, y=1] (s-1)+ \mathbb{I}[s>0, y=0] s}$. Denote the 
	difference of value functions as ${\Delta {V}^\pi_{t}(s)}
	=\tilde{V}^\pi_{t} (s-1) -\tilde{V}^\pi_{t} (s)$, then the full-information $Q_t$ function is $0$ for $t=T$, ${\pyax R(a)\mathbb{I}[s>0]}$ for $t=\horizon-1$, and for $t < \horizon-1$:
	\if \fullpage 1 
	$$\textstyle 
	\;\; 
	Q_t(s,x,a) 		= \pyax
	(
	R(a) \mathbb{I}[s>0] 
	+
	{\Delta {V}^\pi_{t+1}(s)} )
	+   \tilde{V}^\pi_{t+1} (s).  
	$$ 
	\fi 
		\if \fullpage 0
\begin{align*}&\textstyle 
	\;\; 
	Q_t(s,x,a) 		= \\
	&\textstyle\pyax
	(
	R(a) \mathbb{I}[s>0] 
	+
	{\Delta {V}^\pi_{t+1}(s)} )
	+   \tilde{V}^\pi_{t+1} (s).  
\end{align*}
	\fi

\end{example}
Next we describe in words other examples that also fit in the model of  \Cref{diag-stateful-policy-learning} but defer their specific mathematical formulations to the appendix.
\begin{example}[Multi-item network revenue management (informal) \cite{gallego1997multiproduct,gallego2019revenue}]\label{ex-multinrm}
This extends \Cref{ex-singlenrm} with multivariate outcomes (contextual demands for different products). 
	We augment the exogenous context arrival process with  \textit{product arrival types} and product-conditional context distributions.
\end{example}
\begin{example}[Spatial pricing and repositioning (informal, contextual adaptation of \cite{el2020lookahead,bimpikis2019spatial})]\label{ex-spatialpricing}
The state space is
the number of cars at each station in a ridesharing system.
We augment the exogenous information process via uniformized arrivals at a station and origin-destination requests.
The individual contextual response is ride acceptance/rejection at a price; reward is revenue and a lost sales cost. 
\end{example}
\paragraph{Structure that satisfies or does not satisfy assumptions.}
We have discussed classical examples that instantiate these assumptions. However, more complex modeling could violate them. \Cref{asn-X-exogeneous-context} would not be true if customers had full observation of the system and responded to it, such as in queuing if customers can observe queue length and balk. Or, for \cref{ex-spatialpricing}, if customer arrivals are correlated with system state due to unobserved confounders, such as weather patterns that lead to higher propensity to accept a ride and higher customer demand at other locations. In the context of \cref{ex-spatialpricing}, \Cref{asn-lin-mdp-pot-out} is true if the underlying system state $S_t$ (cars at other stations) is not a confounder because it does not affect an \textit{individual's} demand in response to price. However, if system transitions modeled stochastic travel times where system state variables (such as congestion) did causally affect outcomes, \Cref{asn-lin-mdp-pot-out} may not hold.

\section{Related work }\label{sec-relatedwork}
In the main text we only highlight the most closely related work; see \Cref{apx-sec-relatedwork} for further discussion.
\paragraph{Off-policy policy learning for offline sequential decision-making.}
There has been extensive work on off-policy evaluation and learning in the sequential setting. We focus on work that builds on statistical model-free approaches, including doubly robust off-policy evaluation in incorporating value-function control variates \cite{thomas2016data,jl16,zhang2013robust}, and study of the efficient influence function 
\cite{kallus2019double,bibaut2019more,kallus2019efficiently}, as well as MIS or fitted-Q-evaluation \cite{yin2021near,duan2020minimax,le2019batch,hu2021fast}.  

In general, off-policy evaluation in the sequential setting either includes rejection sampling on entire trajectories (even with doubly-robust augmentation) \cite{thomas2016data}, or introduces marginalized density ratios \cite{yin2021near,kallus2019double} which in the finite-horizon setting cannot be optimized in a backwards-recursive fashion or are policy dependent. The latter prevents direct translation of improvements in statistical OPE to off-policy policy optimization except by exhaustive search over the policy class. \cite{nie2020learning} similarly specializes OPE to a different setting, optimal stopping, which admits policy-independent nuisance functions. We similarly develop structure-dependent improvements in dependence on nuisance functions, but for different structure.

Our estimator, derived via the modeling analysis in the next section, does not require rejection sampling on entire trajectories. Therefore we show statefulness is in fact more closely related to single-timestep off-policy evaluation and learning \cite{dell2014,kt15,js15norm,wager17}.
We do not claim novelty relative to the extensively-studied doubly-robust estimation in sequential OPE \cite{jl16,tang2019doubly}: rather we show that specializing to policy structure allows for retaining statistical improvements from double robustness with reduced dependence on nuisance functions (two instead of $T+1$). %

\paragraph{Online contextual decision-making with constraints and algorithmic analysis under known distributions.} 
In the main text we provide an abridged discussion; see \Cref{apx-sec-relatedwork}. Contextual bandits with knapsack (CBwK \cite{badanidiyuru2018bandits}) does consider both contexts and statefulness, relative to an extensive literature (typically model-based) on \textit{either} contextual \cite{cohen2016feature,javanmard2016dynamic,qiang2016dynamic,shah2019semi,ban2020personalized,chen2021statistical} or stateful \citep{huh2011adaptive,besbes2012blind,agrawal2019learning} problems%
.
The closest work is \cite{agrawal2016efficient},
but it considers the Lagrangian relaxation of the resource constraints. (Recall under \Cref{asn-stationary-behavior-policy} we do not consider adaptive data collection; but nonetheless CBwK is a useful comparison.)
Relative to CBwK and pricing bandits, we consider a general MDP embedding and our sample complexity analysis and algorithm do not require specific structure of the reward beyond \cref{asn-X-exogeneous-context,asn-lin-mdp-pot-out}. Our approach is particularly beneficial in handling high-dimensional context variables $X_t$. Naively analyzing approximate linear program arising from state aggregation on $X_t$ incurs statistical bias in general due to discretization. On the other hand, structure-specific analysis of any such problem, such as network revenue management or online packing problems, will generally obtain stronger approximation guarantees for the online setting although we do not comment on direct translation of online algorithms or approximation algorithm-type guarantees, e.g. benchmarked to the fluid relaxations, to the contextual setting. Note \cite{bray2019multisecretary} highlights dependence of recent constant-regret approximation guarantees on discrete distributions, while our setting of contextual responses corresponds to the case of continuous valuations. We defer a comprehensive comparison to future work. Our work is on offline evaluation in such problem settings, and can for example allow off-policy evaluation of policies (when they are not history-adapted) from confounded observational data. 

\section{Off-policy evaluation and learning in the \emph{Marginal MDP}}\label{sec-marg-mdp}
\textbf{Marginal MDP construction.} 
\Cref{problem-setup} described the generating process of the data. We now marginalize over contexts and (policy-induced) outcomes in a lifted \textit{marginal MDP} on a discrete state space and continuous action space where actions are given by policy parameters. 
This MDP is purely a \textit{conceptual device} which is used in the analysis. Direct OPE methods cannot be used in the marginal MDP because observations in the dataset correspond to variation over different \textit{actions}, but not necessarily different \textit{policies} that are actions in the marginal MDP. We develop this construction to justify the use of single-timestep off-policy evaluation, which we denote as $\pypi$:
\begin{equation}\label{eqn-batch-policy-value} 
	{
		\pypi=
		\sum_{a\in \Aa} \E[\pi_e(a\mid X) \pyaX ]}
\end{equation}

To summarize, the marginal MDP is ${\widetilde{\mathcal{M}}=(\Ss,\Pi,\margP, \margR,\horizon)}$, where the action space is the space of ($s$-dependent) policy functions of $x$ and transitions and rewards marginalize over context arrivals.
The key modeling insight is that expectations over individual exogenous arrivals may be estimated via a distribution of iid arrivals; e.g. estimate \cref{eqn-batch-policy-value} by single-timestep off-policy evaluation. 
\begin{description}
	\item The marginal MDP state space is the system state space, $\mathcal S$.
	\item The action space is the set of parametrized policies, $\mathcal A(s) = \Pi(s),$ where $\Pi(s) = \{ \pi(s, \cdot) \in \Pi \}$ is the set of policy functions given $s$.
	\item Transitions between $s$ and $s'(s,y)$ occur with probability $\pypi$, (\cref{eqn-batch-policy-value})
	\item Reward is $
	\textstyle { \margR(s,\pi)   = \sum_{a} 
	\sum_{s',y\mid s}	\E[\pi(a\mid X) 
		 \mathbb{I}[ Y(a) =y] R]
	},
	$ the expected reward induced by context-conditional policy actions and corresponding outcomes, where $R=R(s,a,s'(s,y))$.
\end{description} By construction, policy values and optimal policies are equivalent in the marginal and full-information MDPs (under a policy class that is a product class in $s$). (Note that higher-order moments are not equivalent.)
\begin{proposition}\label{prop-mdp-equivalence} 
Assume the policy class $\Pi$ is a product space over $s\in \Ss, t \in [T]$. The marginal MDP $\widetilde{\mathcal{M}}=(\Ss,\Pi,\margP, \margR,\horizon)$ has the same optimal policy, and policy value $V(s), Q(s)$, as the full-information MDP with policy class $\Pi$ and marginal policy values $\tilde{V}(s), \tilde{Q}(s)$ when ${\mathcal M}=(\Ss\times\Xx\times\Yy, \Aa, P, R,\horizon)$.
\end{proposition}

\textbf{Example: Marginal value function for single-item pricing.} In the marginal MDP for \Cref{ex-singlenrm}, 
\begin{align*}%
\textstyle	
&\margP(s-1 \mid s, \pi) 
= \py{1}\\
&s'(s,y) = \mathbb{I}[s>0] \mathbb{I}[y=1](s-1) \\
&\tildeV_t^{\pi_{t:T}}(s) = 
	\margR(s,\pi)
	+ \tildeV_{t+1}^{\pi_{t+1:\horizon}}(s) + \pypi \Delta V^{\pi_{t+1:\horizon}}_{t+1}(s)
\end{align*}
\textbf{Estimation via fitted value evaluation and iteration in the marginal MDP.} 
We define the propensity score and outcome model as follows: \if \fullpage 1 
$$
\textstyle e_t(a , x) = P(A_t =a\mid X=x),\qquad \mu(y\mid a,x) = P(Y = y \mid A=a,X=x ).
$$
\fi 
\if \fullpage 0
\begin{align*} 
&\textstyle e(a \mid x) = P(A_t =a\mid X=x)\\
&\textstyle \mu(y\mid a,x) = P(Y = y \mid A=a,X=x ).
\end{align*} 
\fi 
The propensity score only controls for $X$: while we allow the underlying behavior policy to be state-dependent, \Cref{asn-lin-mdp-pot-out} implies that adjusting for $X$ is sufficient to estimate the marginalized transition, \cref{eqn-batch-policy-value}, because the state does not affect the outcome. To achieve the orthogonality and rate double-robustness benefits of the doubly-robust estimator we next introduce, we use two-fold sample splitting in trajectories and timesteps. 
We use cross-time fitting and introduce folds that partition trajectories and timesteps $k(i,t)$. For $K=2$ we consider timesteps interleaved by parity (e.g. odd/even timesteps in the same fold). We let $k(i,t)$ denote that nuisance $\hat \mu^{-k(i,t)}$ is learned from $\{ X^{(i)}_{t'},Y^{(i)}_{t'} \}_{i \in \mathcal I_{k(i)}, t' \op{mod} 2 = t \op{mod} 2}$, e.g. from the $-k(i)$ trajectories and from timesteps of the same evenness or oddness but is only used for evaluation in the other fold. Interleaving between timesteps insures downstream policy evaluation errors are independent of errors in nuisance evaluation at time $t$. 

We let $\hpypi$ denote the empirical estimate: we verify that the standard doubly robust estimator for single-timestep offline policy learning, reweighting the empirical transitions in observational data, estimates the transition probabilities in the marginal MDP.
\begin{proposition}[Single-time-step doubly robust estimator of transitions in the marginal MDP.]\label{lemma-batchpolicyest-firststage} Let 
\begin{align}&\Gamma_t^i(y\mid a) \defeq
	\textstyle	\frac{\mathbb{I}[Y_{t}^i=y]-\hat{\mu}^{-k}(y \mid A_t^{i}  X_t^{i})}{\hat{e}_{t}^{-k}\left(A_t^{i}\mid X_t^{i}\right)}
	\mathbb{I}[A_t^i=a]
	+\hat{{\mu}}^{-k }(y \mid a , X_t^i)\nonumber
	\\
&	\hpypi 
		\defeq ({N\horizon})^{-1} \sum_{t=1}^\horizon 	\sum_{i=1}^N  \sum_{a \in \Aa} \pi(a \mid X_t^i)  {\Gamma}_t^i(y\mid a).     
\end{align} 
$
\hpypi
$ is an unbiased estimator of 
$
\pypi
$ if at least one of $\hat\mu$ or $\hat e$ are unbiased. 
\end{proposition}
\Cref{lemma-batchpolicyest-firststage} verifies \textit{orthogonality}, that the estimator is doubly-robust against misspecification of one of $\mu$ or $e$. 	The estimator only adjusts for contexts because \Cref{asn-lin-mdp-pot-out} specifies that the state variable is not a confounder. \Cref{lemma-batchpolicyest-firststage} considers the stationary case; when $X_t$ is time-varying but non-adversarial with fixed distributions, similar data-pooling is possible by estimating density ratios.\footnote{	In revenue-management settings, it is common for $X_t$ arrivals to be nonstationary. While online algorithms considers adversarial arrival distributions, relevant arrivals may also have highly structured nonstationarity, e.g., ``business-class" arrivals arriving later on. To a limited extent, \textit{adversarial} arrivals could also be modeled by robustness, e.g., using the approach of \cite{kallus2020minimax} over density ratios for each timestep's subproblem in \Cref{alg-backwards-recursion}. }

Given a generic estimator $\hpypi$ for the marginal transition probability $
\pypi
$, we can construct a $Q$-estimate as follows: $\hpypi$ can be the doubly robust estimator as in \Cref{lemma-batchpolicyest-firststage} or alternatively the IPW estimator (simply let $\hat\mu^{k(i,t)}=0$ in \Cref{lemma-batchpolicyest-firststage}) or direct method estimator (simply let $\hat e^{k(i,t)}=\infty$ in \Cref{lemma-batchpolicyest-firststage}). We use backwards recursion to evaluate 
$\htPiV{{\tpi}_{t:\horizon}}_{t}(s)$ using model-based evaluation with $\hpypi$ in the marginal MDP. 
\if \fullpage 1
\begin{equation}\label{eqn-policy-value-estimation}
\textstyle	\htPiQ{\pi, {\tpi}_{t+1:\horizon}}_t(s,\pi) = \sum_{(s',y)\mid s} 
\hpypi
\left(R(s,a,s')+ \htPiV{{\tpi}_{t+1:\horizon}}_{t+1}(s') \right) 
\end{equation} 
\fi 
\if \fullpage 0 
\begin{align}\label{eqn-policy-value-estimation}
	\textstyle	&\htPiQ{\pi, {\tpi}_{t+1:\horizon}}_t(s,\pi)= \nonumber
	\\
	& \sum_{(s',y)\mid s} 
	\hpypi
	\left(R(s,a,s')+ \htPiV{{\tpi}_{t+1:\horizon}}_{t+1}(s') \right) .
\end{align} 
\fi 
\textbf{Policy Learning.} 
When the policy space is in fact a product set over the state space (i.e., the policy being optimized can vary independently for every value of the state), we study a policy learning proposal in \Cref{alg-backwards-recursion} which implements backwards-recursive policy learning (which can be understood as fitted value iteration in the marginal MDP) to determine the optimal policy vector $\tpi$. 

\begin{algorithm}[t!]
	\caption{Backwards-Recursive Policy Learning}\label{alg-backwards-recursion}
	\begin{algorithmic}[1]
		\State Input: estimate $\hpypi$, policy class $\Pi_{0:T}$
		\For{$t = \horizon, \dots, 0$}: 
		\For{$s \in \Ss$}: 
		\State Estimate off-policy value $\htPiQ{\pi , {\hat{\tpi}}^*_{t+1:T}}_t(s,\pi)$ via \cref{eqn-policy-value-estimation}
		\State Optimize $\hat{\pi}^*_{t,s} \in \underset{\pi \in \Pi_t(s)}{\arg\max}  \;\;
		\htPiQ{\pi,{\hat{\tpi}}^*_{t+1:T}}_t(s,\pi)  $ and update $\htPiV{{\hat{\tpi}}^*_{t:T}}(s) \gets \htPiQ{{\hat{\tpi}}^*_{t:T}}_t(s,\hat{\pi}^*_{t,s} ) $

		\EndFor
		\EndFor
		\State
		\Return { $\hat{\tpi}^* = \{ \hat\pi_{t,s}^* \colon t \in [T], s \in \Ss \} $ }
	\end{algorithmic}
\end{algorithm}
\section{Analysis}\label{sec-analysis}
\textbf{Sample complexity.} We first provide a generalization bound for \Cref{alg-backwards-recursion} on the out-of-sample regret $\tildeV^{\hat\pi^*}_0$, the true value achieved by the sample-optimal policy $\hat\pi^*$, relative to the best-in-class policy, $\tildeV^{\pi^*}_0$.
We assume the policy class at a given $s,t$ has restricted functional complexity in the sense of a finite entropy integral of the covering numbers \cite{van1996weak,wainwright2019high}. In the main text we use the VC dimension $\vc$ for binary actions; in the appendix we include corresponding statements for multi-class notions such as Natarajan dimension \cite{mohri2018foundations}.

\begin{theorem}[Sample complexity and rate double-robustness ]\label{thm-policy-learning-sample-complexity-uniform-convergence} 
	 Suppose $\nu^{-1} \leq {e(a\mid x)\leq 1-\nu^{-1}}$ uniformly over $a,x$, for $\nu>0$, (overlap) and for some rates $ 0 < r_1, r_2 < 1$ and constants $C_1,C_2$,
	  we have uniformly consistent estimation of nuisance estimates 
	 \if \fullpage 1 
	 $$
	 \textstyle
	 {\E[ (\mu(y\mid a,X)-\hat\mu(y\mid a,X))^2] = o_p( {n^{-r_1}} ) 
	 },  \;\;
	 {\E[ (e(a\mid X)-\hat e(a\mid X))^2] = o_p({n^{-r_2}}) },$$
	 \fi
	 \if \fullpage 0 
\begin{align*} 
	 \textstyle
	 &{\E[ (\mu(y\mid a,X)-\hat\mu(y\mid a,X))^2] = o_p( {n^{-r_1}} ) 
	 },\\  \;\;
	&{\E[ (e(a\mid X)-\hat e(a\mid X))^2] = o_p({n^{-r_2}}) }
	 \end{align*} 
	 \fi 
	 where $r_1 + r_2 \geq 1.$ Then there exists a random variable $\kappa =o_p(({NT})^{- \frac 12})$ so that w.p. $\geq 1-\delta$,
$$\tildeV^{\pi^*}_0 - \tildeV^{\hat\pi^*}_0 	\leq \frac{ 5\nu  \Rmax  ( \horizon^{1/2}  + \nicefrac{1}{2}\horizon^{3/2})   \sqrt{
		\vc\log(\frac{5 \horizon \vert \Yy \vert }{\delta})} }{\sqrt{N } }  + \kappa.$$
\end{theorem}

The $\kappa$ term arises because we decompose the value difference with an \textit{oracle} estimator using the true nuisance functions $\mu,e$, and obtain a high-probability bound on the leading order term. The final bound is of order $O_p(N^{-\frac 12}T^{\frac 32})$. The proof follows standard techniques, combining single time-step uniform convergence with the performance difference lemma. However, it is the previous modeling analysis and our derived estimator that permits this reduction.

The main improvement in \Cref{thm-policy-learning-sample-complexity-uniform-convergence} is in specializing to statefulness so only two nuisance functions are required, rather than $\horizon\times\abs{\Pi_{0:T}}$ many as would arise in the case of $Q$-function nuisances. In appendix \cref{apx-sec-discussion} we also discuss improvements in dependence on concentratability coefficients/sequential overlap.

Finally, we verify the nuisance rates are as achievable from pooled episodes as they would be from iid data. In the main text, $\lesssim$ omits polylogarithmic factors; see appendix \cref{apx-sec-nuisancerates} for a full statement. This is a direct corollary of 
\cite{bojun2020steady} 
and a convergence rate from mixing data given in \cite[Thm. 5]{farahmand2012regularized}; $\alpha$ describes a
mild capacity assumption on the covering numbers of nonparametric nuisance function classes.
\begin{corollary}[Estimation of nuisance functions]\label{cor-nuisance-rates}
	Suppose nuisance function estimator $\hat e(a\mid x)$ is learned from pooled-episode data $\{ (X_t^i, A_t^i, Y_t^i)\}^n_{0:\horizon},$  
	and ${\mu(y\mid a, x)}$ from $\{ (X_t^i, A_t^i)\}^{n}_{0:\horizon}$ by regularized least squares. Assume that $A,Y$ are bounded a.s., and well-specification of $e,\mu$. Assume there exist $C>0,0\leq \alpha < 1$ such that for any $\epsilon, R>0$, the covering numbers are bounded, $\mathcal{N}_2(\epsilon, \mathcal{B}_R, X_{1:N})\leq C(R/\epsilon)^{2\alpha}$. Then there exists $c_1, c_2>0$ such that for large enough $n$, for any fixed $\delta \in (0,1)$, with probability $\geq 1-\delta$,
\begin{align*} 
	&\E[(e(a\mid X) - \hat e(a \mid X))^2] \lesssim  n^{-\frac{1}{1+\alpha}},\\
	&\E[(\mu(a\mid X) - \hat \mu(a\mid X))^2] \lesssim  n^{-\frac{1}{1+\alpha}}.
\end{align*} 
\end{corollary}
\Cref{cor-nuisance-rates} verifies minimax rate-optimality of estimation in this setting \cite{farahmand2012regularized,yang1999information}, e.g. estimating the nuisance functions from pooled episodes achieves the minimax nonparametric rate that prevails in the absence of dependence up to polylogarithmic terms.
\begin{remark}[Extension to continuous states]
We can extend to doubly-robust policy evaluation with continuous transitions by invoking recent advances in estimating counterfactual distributions 
(\cite{kennedy2021semiparametric}) but further work is required for function approximation for policy learning. See {\Cref{apx-sec-cont-transition}}.
\end{remark}
\textbf{Structural analysis: error propagation  in dynamic pricing.}\label{secsub-structuralanalysis} 
We study the possible drawbacks of naively plugging-in a confounded model (DM) by analyzing the sequential error propagation. We provide structural conditions for error \textit{persistence} in the sequential setting:
error 
could ``persist" if model error in transitions and value functions continues to impact downstream learned policies or ``attenuate" if these cancel out in the sequential setting.
We specialize to \Cref{ex-singlenrm}:
similar results should hold for other settings with less interpretable sufficient conditions. Define the threshold $\theta^*$, true conditional expectation ratio $\outcomeratio^*$, and optimal (threshold) policy $\pi^*$:
 \begin{align}\label{eqn-structural-analysis-thresholds}
&\textstyle
\theta^*_t(
s,\Delta V
)=\frac{R(0) + \Delta V
}{R(1) + \Delta V
}, 
\;\; \outcomeratio^*(x) =	\frac{\mu(1\mid 1,x)}{\mu(1\mid 0,x)}, \;\; \\
&\textstyle
\pi^*(1 \mid s,x) = \mathbb{I}
\bracks{
	\outcomeratio^*(x) >\theta^*_t
}. 
\end{align}
The confounded-optimal policy incurs error from $\Delta\hat{\mu}$ or $\Delta V^{\hat\pi^*}$. We reparametrize the decision boundary on $\Delta\hat{\mu}$ relative to $\theta^*,\Delta\mu^*$.
Then the biased threshold $\hat\theta^*$ is related to the true $\theta^*$ by the  pointwise error $\delta(a,x)$: 
\begin{align}\delta(a,x) &=\hat\mu(y\mid a,x)- \mu(y\mid a,x)
,\\ \;\;
\hat\theta^*&={\theta^*_t(
	s
	) \cdot \left( 1 +\nicefrac{\delta(0,x)}{ \etaax{0}}\right) - \delta(1,x)}.
\label{eqn-structural-analysis-biased-threshold}
\end{align} 
When self-evident we omit dependence of $\theta^*$ on arguments that remain fixed for brevity. 
Error ``persists" if
the error at different timesteps, including from value estimation, persists in the same direction relative to the optimal policy. We provide a sufficient condition to conclude the direction of error induced from downstream errors in value estimation. In the main text we state a special case; the appendix includes the full theorem for $t<T-2$ with less interpretable conditions.
\begin{theorem}[Conditions for error persistence]\label{thm-example-bias-amplification}
For $t=T-2$, assume $R(1)>R(0)$, without loss of generality.
Then, for $s>2$, 
\if \fullpage 0
\begin{align*}
	&\textstyle	\E[ -\tau(X)
	\mathbb{I} [{\hat{\theta}^*_{T-1}\leq \outcomeratio^*(X) \leq \theta^*_{T-1}}]]\geq0 \\
	&\implies		{\textstyle \hat\theta_{T-2}^*(\Delta V^{\hpi{T-1:T}}_{T-1})< \hat\theta_{T-2}^*(\Delta V^{\spi{T-1:T}}_{T-1}) }.\end{align*}
\fi 
\if \fullpage 1
\begin{equation*}
	\textstyle	\E[ -\tau(X)
	\mathbb{I} [{\hat{\theta}^*_{T-1}\leq \outcomeratio^*(X) \leq \theta^*_{T-1}}]]\geq0 \implies		{\textstyle \hat\theta_{T-2}^*(\Delta V^{\hpi{T-1:T}}_{T-1})< \hat\theta_{T-2}^*(\Delta V^{\spi{T-1:T}}_{T-1}) }.\end{equation*}
\fi 
\end{theorem}

We discuss implications of \Cref{thm-example-bias-amplification} for bias persistence in the context of \Cref{ex-singlenrm}.
\begin{example}[Error persistence in \Cref{ex-singlenrm}.]\label{ex-errorpersistence}
	Suppose 
	$\delta(1,x) >0> \delta(0,x)$ uniformly over $x$, for example if historical price increases were targeted towards those more likely to purchase them 
	and discounts were targeted to those less likely to purchase overall. Then for any $s,\Delta V$, $\hat\theta^*(s,\Delta V) \leq \theta^*(s,\Delta V)$. 
	By assumption on, e.g. price elasticity so that $\tau(x)\leq0$, we expect the sufficient condition of \Cref{thm-example-bias-amplification} to be true so that bias persists; for $s>2$,
\begin{equation*} 
	\hat\theta_{T-2}^*(\Delta V^{\hpi{T-1:T}}_{T-1})< \hat\theta_{T-2}^*(\Delta V^{\spi{T-1:T}}_{T-1})< \theta^*_{T-2}(\Delta V^{\spi{T-1:T}}_{T-1}). 
	\end{equation*}
\end{example}
\section{Empirics}
\begin{figure*}[t!]
	\begin{subfigure}{0.635\textwidth}
		\includegraphics[width=0.5\textwidth]{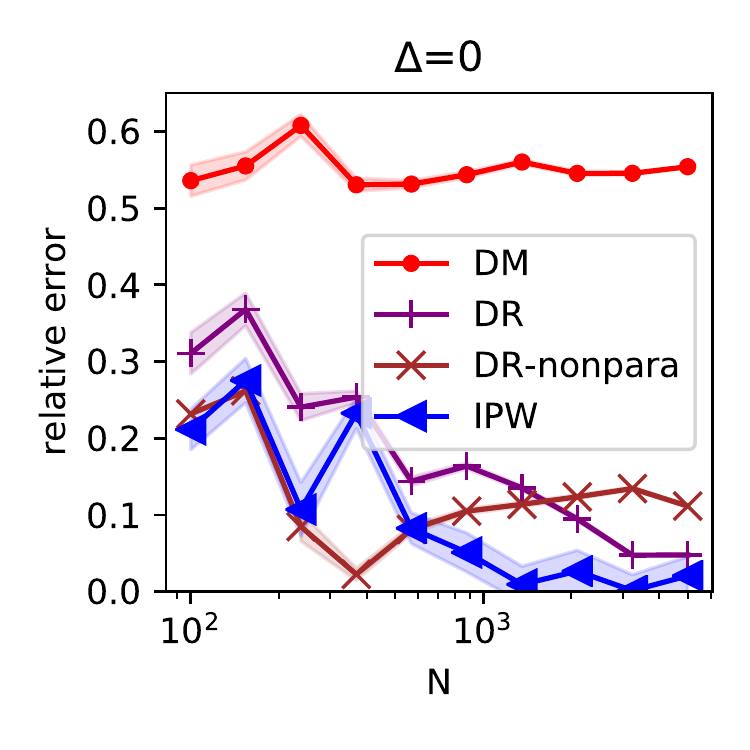}\includegraphics[width=0.5\textwidth]{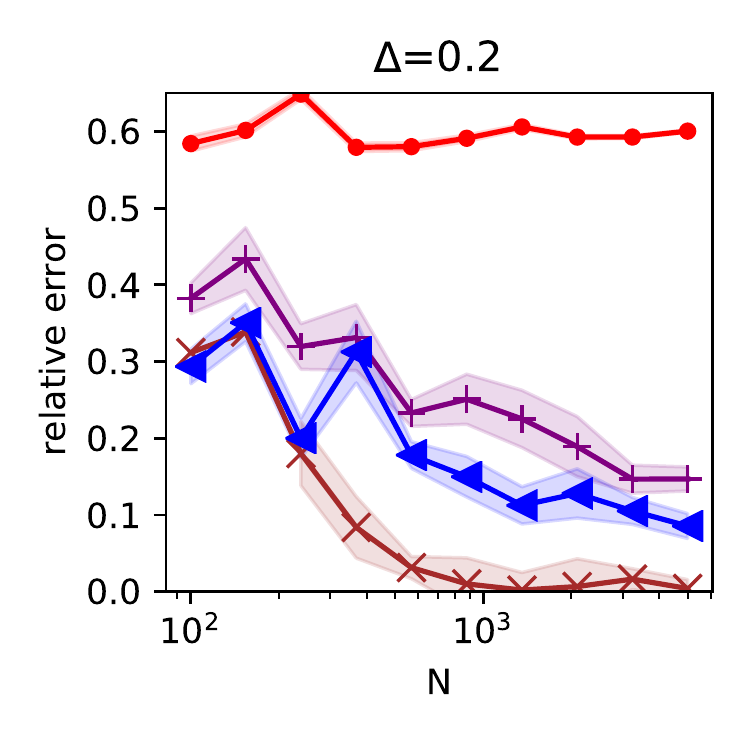}
		\caption{Relative absolute error (y-axis) of off-policy evaluation with increasing  model misspecification ($\Delta$). Lower is better.}	\label{fig-poleval}
	\end{subfigure}\begin{subfigure}{0.025\textwidth}\mbox{}
\end{subfigure}
	\begin{subfigure}{0.40\textwidth}
		\centering
		\includegraphics[width=\textwidth]{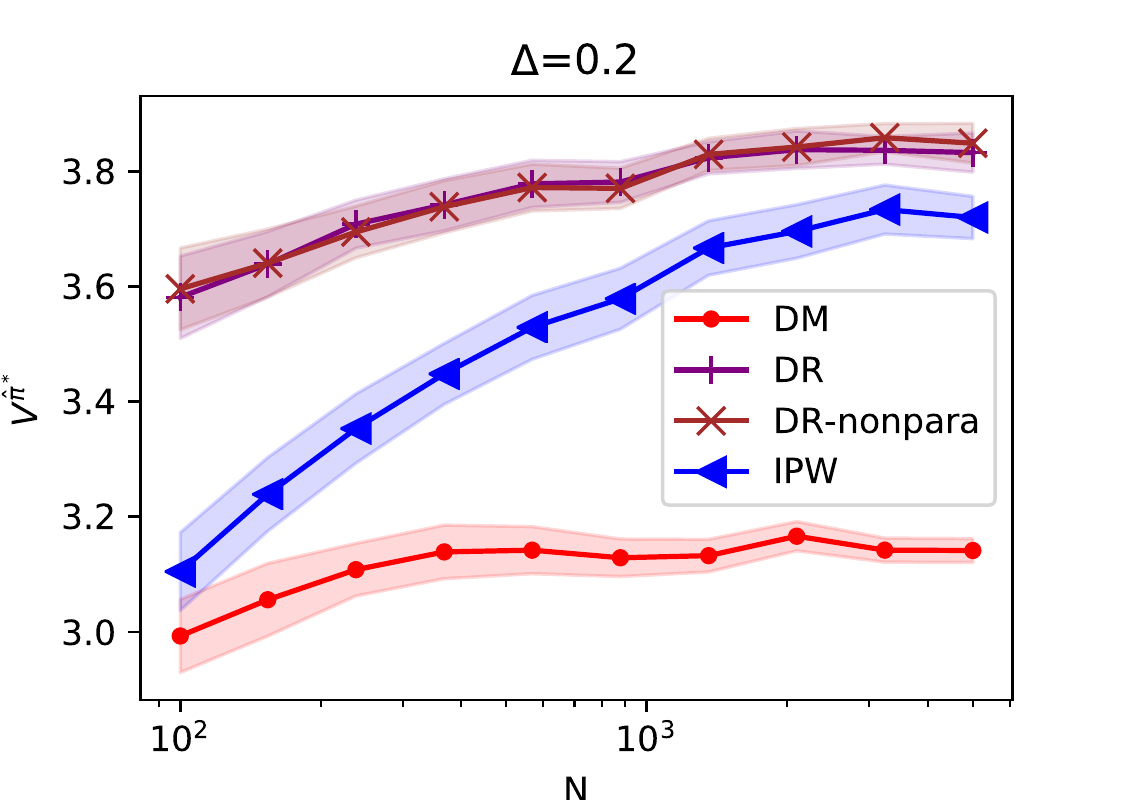}
		\caption{Out-of-sample value $V^{\hat\pi^*}$ (y-axis) of policy optimization: threshold policies optimized via \Cref{alg-backwards-recursion}. Higher is better.}
		\label{fig-polopt}
	\end{subfigure}
	\caption{Policy evaluation and optimization as more trajectories (x-axis) are collected of $\horizon$-horizon selling in contextual and capacitated dynamic pricing, \cref{ex-singlenrm}; specification of \cref{eqn-dgp}.}
	\label{fig-ope}
		\begin{subfigure}[m]{0.3\textwidth} 
		\centering
	\includegraphics[width=\textwidth]{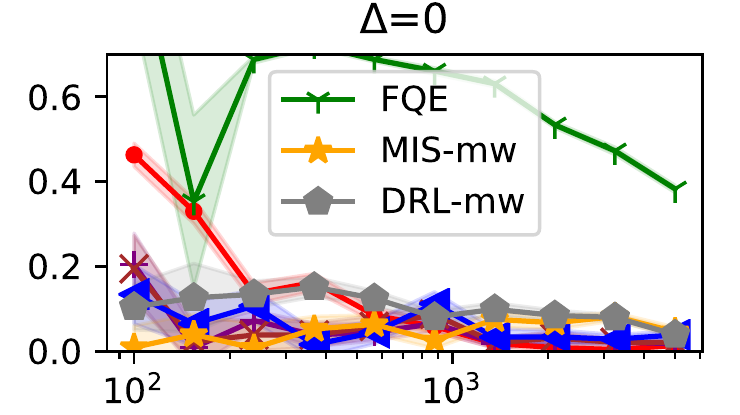}
	\includegraphics[width=\textwidth]{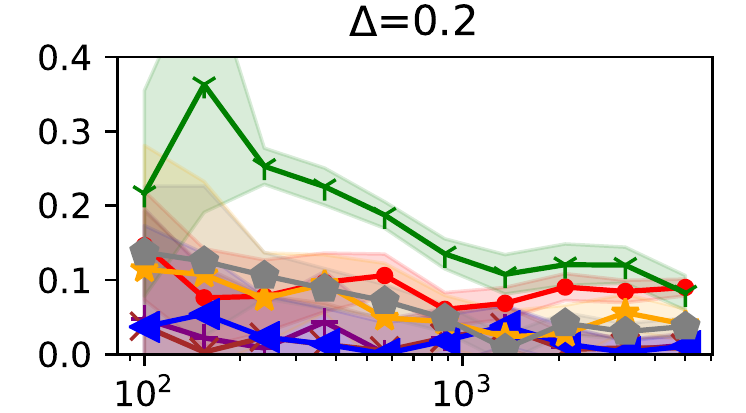}
	\caption{$\Delta =0.2$; further OPE comparison on a different DGP. }
	\label{fig-ope-baselines}
\end{subfigure}\begin{subfigure}[m]{0.3\textwidth} 
		\centering
		\includegraphics[width=0.8\textwidth]{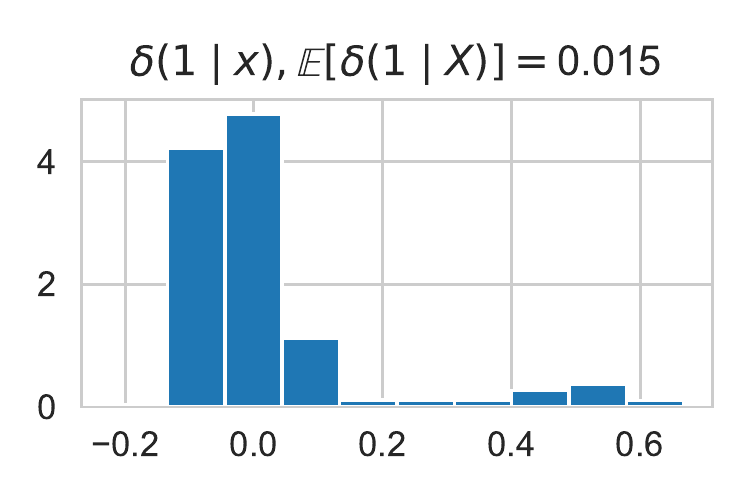}
		\caption{(Normalized) histogram of conditional bias $\delta(1,x)$.}
		\label{fig-structural-analysis-hist} 
	\end{subfigure}\begin{subfigure}{0.05\textwidth}\mbox{}
\end{subfigure}\begin{subfigure}[m]{0.4\textwidth} 
		\centering
		\includegraphics[width=\textwidth]{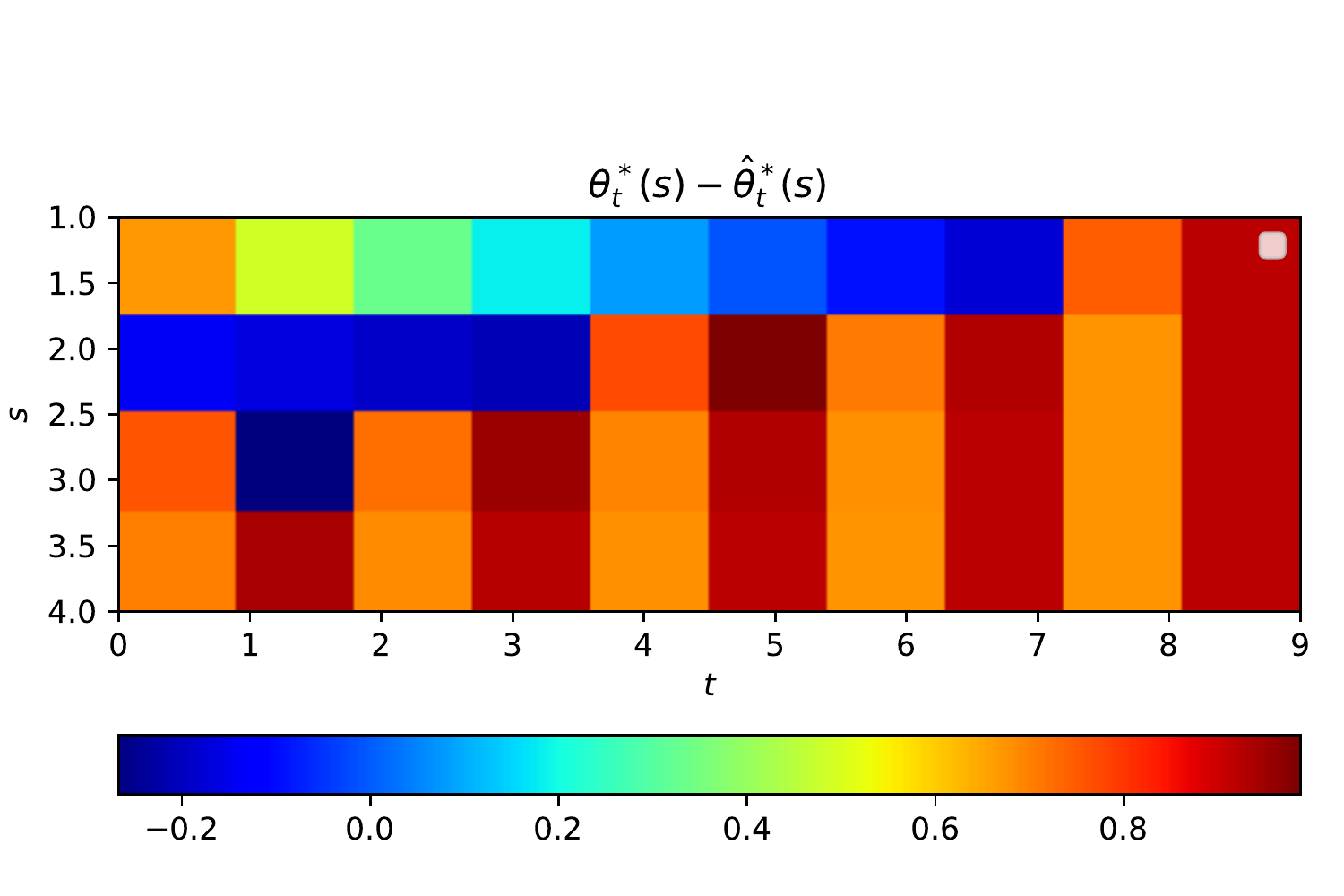}
		\caption{Heatmap of thresholds optimized via DR vs. (biased) thresholds, over states (y-axis) and timesteps (x-axis). Red is direction of single-timestep error (persistence).  }
		\label{fig-heatmap-bias} 
	\end{subfigure}
	\caption{Conditions of \Cref{thm-example-bias-amplification} for error persistence.}\label{fig-structuralconditions}
\vspace{-1\baselineskip}\end{figure*}
\textbf{Data-generating process.} 
We consider a simple example based on single-product dynamic pricing (\cref{ex-singlenrm}), with a response model that is a $\Delta$-weighted mixture model of a logistic specification and a nonlinear specification, where $\sigma(\beta^\top x)= (1+\exp(-\beta^\top x))^{-1}$.
\if \fullpage 0 
\begin{align}
\textstyle 
\mu(1\mid a,x) &= (1-\Delta) \sigma(\beta^\top x + \beta_0 
 p_a )
+ \Delta\sigma(x_0^2  p_a ), \\\;
e_t(1\mid x)&={\sigma(-0.8\beta^\top x )} 
\label{eqn-dgp} 
\end{align}
\fi 
\if \fullpage 1
\begin{align}
	\textstyle 
	\mu(1\mid a,x) = (1-\Delta) \sigma(\beta^\top x + \beta_0 
	\cdot p(a) )
	+ \Delta\sigma(x_0^2 
	\cdot p(a) ), \;
	e(1\mid x)={\sigma(-0.8\beta^\top x )} \label{eqn-dgp} 
\end{align}
\fi
We generate the data corresponding to the outcome specification for parameters 
$ \beta=[-0.75, 0.75]$, $\beta_0 =-2$. 
We learn outcome models $\hat \mu$ by either logistic regression (for $\op{DM}$, direct method or $\op{DR}$, doubly robust) or a neural network for a nonparametric nuisance estimate ($\op{DR-nonpara}$), and the behavior policy by (well-specified) logistic regression. We consider a (time and state-stationary) evaluation policy ${\pi_e(1\mid x) = {\sigma(0.25(\beta^\top x  ))}}$. The time horizon is 10 timesteps, with initial state capacity $s_0=4$.

\textbf{Policy evaluation and optimization.} 
In \Cref{fig-ope} we generate different outcome models with increasing levels of misspecification $\Delta$, evaluate $V_0^{\pi_e}(s_0)$ by Monte Carlo rollouts with $N=10000$ trajectories. We compare $\op{DM}$ with logistic regression nuisance, 
$\op{DR}$ doubly-robust with logistic regression, $\op{DR-nonpara}$ with nonparametric nuisance, and $\op{IPW}$, inverse propensity weighting. (See \Cref{apx-sec-experiments} for further comparison including other baselines and policy optimization in the well-specified $\Delta=0$ case, where the variance drawbacks of $\op{IPW}$ do worse than model-based approaches.) \Cref{fig-poleval} considers off policy evaluation, with absolute relative error on the y-axis and trajectory size on the x-axis (log grid from $N=50,\dots 5000$). When $\Delta=0.2$ the logistic outcome model is misspecified, but orthogonality and the well-specified propensity score ensures estimates are asymptotically unbiased. Similar to other DR settings, although incorporating the outcome model reduces variance, incorporating a misspecified outcome model does worse than just using well-specified $\op{IPW}$, but we see faster convergence from the flexible, nonparametric nuisance which outperforms well-specified $\op{IPW}$. We also compared to nonparametric baselines FQE \cite{le2019batch}, and modified stateful versions of MIS \cite{yin2021near} and DRL \cite{kallus2019double}. However, in this simple setting, the highly flexible nuisance estimators overfit and fail (incurring 40-50\% absolute error). We discuss these baselines in greater detail in ``OPE comparison" in a more favorable data-generating process. 

We then consider policy optimization in \Cref{fig-polopt}, with a rich policy class to avoid misspecification error issues. Motivated by 
\cref{eqn-structural-analysis-biased-threshold}, observe that the optimal threshold policy on the true $\Delta\mu$ is an affine transform relative to a threshold on the estimated $\Delta\hat\mu$ (possibly misspecified, hence biased), with an $x$-conditional term for the conditional error. We approximate optimizing over policies $\mathbb{I}[\Delta\mu>\theta]$ by ranging over all thresholds on $\mathbb{I}[\Delta{\hat\mu}>\theta']$; this approximates the $x$-conditional error term of \cref{eqn-structural-analysis-biased-threshold} by a constant. This is similar to a contextual version of ``bid-price" policies \cite{gallego2019revenue}. We optimize over the class of threshold policies on $\Delta{\hat\mu}$ by enumerating thresholds and evaluating via the estimate from \Cref{lemma-batchpolicyest-firststage}, so the functional specification does depend on the (unadjusted) nuisance estimation. (Therefore the VC dimension of this depends on the VC dimension of the outcome nuisance). The $y$-axis depicts out of sample value (higher is better) averaged over 48 replications. Both $\op{DR}$ and $\op{IPW}$ (inverse propensity-weighted) estimates translate to improvements in optimized policy value. We see dramatic benefits of $\DR$ when $\Delta=0.2$. For small dataset sizes, $\IPW$ suffers from high variance as expected.
Therefore, $\DR$ and its variance reduction estimates achieve sizeable improvements for small amounts of data. As the amount of data grows larger, the performance of $\IPW$ nears that of $\DR$ asymptotically. The DM plug-in approach remains biased and achieves worse performance, even asymptotically. 

\textbf{OPE comparison. }
We compare to state-of-the-art OPE: $\FQE$ of \cite{le2019batch} which does not use the ``stateful" structure, and we also derive ``strong baselines" that leverage some of the structure ($\MIS$ \cite{yin2021near}, $\DRL$ \cite{kallus2019efficiently}). (We reiterate our core contribution is not in general off-policy evaluation but in deriving improvements for this specific structure).
For example, observe that since $x$ is exogenously generated the finite-horizon \textit{state-action density ratio} 
is \textit{independent of $x$}. We endow $\MIS$ and $\DRL$ with this structural information
 (see appendix \Cref{apx-sec-experiments} for details). 
As the general OPE literature prescribes, we use nonparametric nuisances, e.g. multi-layer perceptron with $\op{scikit-learn}$ defaults for all nuisance predictors. We consider a more favorable DGP for OPE comparison in \Cref{fig-ope-baselines}, using \cref{eqn-dgp} with $p=5$ and $\beta=[-0.53 , -0.56, -0.10,  0.40,  0.74], \beta_0 =- 2.39$. 
$\MIS$ does well overall, although empirically we find other DGPs where $\MIS$ underperforms $\FQE$. In the misspecified setting, our doubly robust estimators outperform $\MIS$. $\FQE$, which fits next time-step $Q(s,x,a),$ 
appears to converge but much slower than our approaches. The gap between $\FQE$ and $\DM$ for the (slightly misspecified) case precisely illustrates the benefits of encoding problem structure in \Cref{eqn-qfunction}. 

\textbf{Assessing the structural conditions of \Cref{thm-example-bias-amplification} in practice.}
In \Cref{fig-structuralconditions} we investigate 
assumptions made in
\Cref{ex-errorpersistence} 
(e.g. uniformity of error direction $\delta(1,x)$)
that do not hold exactly. 
\Cref{fig-structural-analysis-hist} plots 
$\delta$: although it is symmetrically distributed for most $x$, there is overall marginal error in the expected direction%
. In the empirical example we optimize over marginal thresholds and so we expect, marginalizing over x, the directional error condition is satisfied. In \Cref{fig-heatmap-bias}, we show a heatmap of $\theta_t^*(s) - \hat{\theta}_t(s)$ over timesteps and state values. As the analysis suggests, for $s >2$ for most timesteps the error persists: red indicates regions where naive thresholds are in the same direction, relative to the optimal threshold, and hence the error persists rather than attenuates over time. 
\section{Conclusion} 
By studying the causal structure of practically relevant problems in operations, we developed specialized off policy evaluation and optimization which demonstrate
the offline version of such problems is easier than a generic MDP. We show analytically that confounding matters, and verify our approach, reducing from $T+1$ nuisances to $2$ and estimating the expectation of a transition via the expectation over a population, achieves practical benefits. 
Future directions include function approximation in $s$ and generalizing to other hierarchical structure.

\bibliographystyle{abbrvnat}
\bibliography{opt-under-uncertainty,prediction-and-opt,library,sensitivity-and-rl,sensitivity_and_rl,stateful-ope}
	
	\clearpage
\addcontentsline{toc}{section}{Appendices}
\setcounter{section}{0}
\renewcommand{\thesection}{\Alph{section}}

\onecolumn
\paragraph{Outline of appendix}

\begin{description}
	\item \Cref{apx-sec-marginalmdpintro} analyzes the marginal MDP construction of \Cref{sec-marg-mdp}. 
	\item \Cref{apx-sec-marginalmdp-policylearning} provides analysis of the main estimation strategy and policy learning sample complexity analysis, e.g. proofs of \Cref{sec-analysis}.
	\item \Cref{apx-sec-discussion} provides additional discussion.
	\item \Cref{apx-sec-experiments} includes additional empirics and discussion of computational details. 
\end{description}

\section{Proofs for \Cref{sec-marg-mdp}}\label{apx-sec-marginalmdpintro}

\begin{proof}[Proof of \Cref{prop-mdp-equivalence}]\label{proof-mdp-equivalence} 
	We argue via backwards induction, though the proof follows largely from construction of the marginal MDP. Consider the last timestep, $t=T$: equivalence follows by construction. Next consider the inductive step. The inductive hypothesis posits equivalence of policy values $V_{t+1}(s)$ in the marginal MDP and $\tilde{V}_{t+1}(s)$, the marginalized policy value in the full-information MDP, as well as equivalence of optimal policies so that $V_{t+1}(s)=\tilde{V}_{t+1}(s)$.

	Recall that each policy $\pi \in \Pi$ in $\mathcal M$ corresponds to an action $\pi \in \Pi$ of $\widetilde{\mathcal M}$ with transition probability: 
	\begin{equation}\label{eq-mdp-equivalence}
		\margP(s'\mid s,\pi) = \pypi \defeq \int\int P(s'\mid s,a,x)\pi(a\mid x) f(x)dx  \end{equation}
	Equivalence of the policy values follows from backwards induction of expected rewards and definition of the marginal MDPs. Equivalence of optimal policies follows from equivalence of first moments and equivalence of the policy classes. Equivalence of the inductive step follows from equivalence of policy classes, identically distributed arrivals $X_i$ within a timestep . 
\end{proof}

	\begin{proof}[Proof of \Cref{lemma-batchpolicyest-firststage}]

		Double robustness is standard and follows standard arguments in the single time-step policy learning literature \cite{dell2014,wager17}. The claim follows by observing that by the restricted causal structure \Cref{asn-lin-mdp-pot-out}, $Y(a) \indep S \mid X$, so it is sufficient to adjust for IPW weights ${P(A=a\mid X =x)}$. 
		
		For completeness, we verify double-robust unbiasedness properties of $\hat V_{t}(s)$. 
		The proof follows by backward induction. The fact that $V_{T+1}(s) = 0$ and estimation is unbiased follows from double-robustness in the single-timestep setting for $\hat V_T(s)$ for all $s \in \Ss$. 
		
		Next we show the inductive step. Suppose $\hat V_{t+1}(s)$ is unbiased, for all $s \in \Ss$. We require that the evaluation error is independent $( \hat V_{t+1}^{\text{tDR}}(s')-V_{t+1}(s') )$ from the nuisance evaluation error, which can be easily satisfied by appropriate sample splitting.

		If $\mu$ is unbiased: 
		\begin{align} 
			\E[ \hatV_t(s) ]	&
			=
				 \E
				\left[\sum_{s',y\mid s} 
				\sum_a
				\pi_e(a\mid x) \mu_t(y\mid a, x)
				(R(y) + \hat V_{t+1}^{\text{tDR}}(s'))   \right] 
		\label{eqn-dr-muunbiased-1}
		\\
			& \qquad
			+ 
			\E\left[\sum_{s',y\mid s} 
			\sum_a
			\pi_e(a\mid x)
				\left(
				\frac{ \indic{A_t=a}}{\pi_b(a\mid x)}\left(\indic{Y=y} -\mu_t(y\mid a, x) \right) 
				\right)
			(R(y) +  V_{t+1}(s'))   \right] 		\label{eqn-dr-muunbiased-2}
			\\
			& \qquad
			+ 
			\E\left[\sum_{s',y\mid s} 
			\sum_a
			\pi_e(a\mid x)
				\left(
				\frac{ \indic{A_t=a}}{\pi_b(a\mid x)}\left(\indic{Y=y} -\mu_t(y\mid a, x) \right) 
				\right)
					( \hat V_{t+1}^{\text{tDR}}(s')-V_{t+1}(s') )  
		\right] 	\label{eqn-dr-muunbiased-3}
		\end{align} 
		Note \cref{eqn-dr-muunbiased-1} $=V_t(s)$ by well-specification,  \cref{eqn-dr-muunbiased-2} $=0$ by unbiasedness of $\mu_t$ and cross-fitting so the estimation errors are independent, and the first term of \cref{eqn-dr-muunbiased-3} is expectation-0 by the previous argument and $\E[( \hat V_{t+1}^{\text{tDR}}(s')-V_{t+1}(s') )  ]=0$  by the induction hypothesis and cross-fitting. 
		
		If $e$ is unbiased: 
		
		\begin{align*}
			&\E[ \hatV_t(s) ] \\
			&= 
			\E\left[ \sum_{s',y\mid s} 
			\sum_a
			\pi_e(a\mid x)
			\left(
			\underbrace{	
				\frac{ \indic{A_t=a}}{\pi_b(a\mid x)}\indic{Y=y}  
			}_{\text{unbiased IPW estimator}}
			+ \mu_t(y\mid a, x) 
			\underbrace{	\left( 1 - 	\frac{ \indic{A_t=a}}{\pi_b(a\mid x)}\right)
			}_{= 0 \text{ by unbiasedness of }  e }	\right)
			(R(y)+ V_{t+1}(s')) 
			\right]\\
			&+\E\left[ \sum_{s',y\mid s} 
			\sum_a
			\pi_e(a\mid x)
			\left(
			\mu_t(y\mid a, x) 
			\underbrace{	\left( 1 - 	\frac{ \indic{A_t=a}}{\pi_b(a\mid x)}\right)
			}_{= 0 \text{ by unbiasedness of }  e }	\right)
			\underbrace{(\hat V_{t+1}^{\text{tDR}}(s')  -V_{t+1}(s'))	}_{ = 0 \text{ by inductive hypothesis }}
			\right]
		\end{align*}

	\end{proof}
	\section{Proofs for \Cref{sec-analysis}}\label{apx-sec-marginalmdp-policylearning} 
	\subsection{Sample complexity proofs}
	\subsubsection{Concentration preliminaries} 
We introduce the uniform convergence setup we use to provide tail inequalities. We will apply a standard chaining argument with Orlicz norms and introduce some notations from standard references, e.g. \cite{vershynin2018high,pollard1990empirical,wainwright2019high}. A function $\phi: [ 0,\infty) \to [ 0,\infty)$ is an Orlicz function if $\phi$ is convex, increasing, and satisfies $\phi(0) = 0, \phi(x) \to \infty$ as $x\to \infty$. For a given Orlicz function $\phi$, the Orlicz norm of a random variable $X$ is defined as $ \norm{ X }_\phi = \inf \{ t> 0 \colon \E[ \Phi(\norm{X}\mid t ) ] \leq 1 \}  $. The Orlicz norm $\norm{Z}_\Phi$ of random variable $Z$ is defined by $\norm{Z}_\Phi = \inf \{  C > 0 \colon \E[ \Phi( Z / C) ] \leq 1  \}. $
A constant bound on $\norm{Z}_\Phi$ constrains the rate of decrease for the tail probabilities by the inequality $ \pr( \vert Z \vert \geq t ) \leq  1 /\Phi(t/C) $ if $C = \norm{Z}_\Phi$. For example, choosing the Orlicz function $\Phi(t) = \frac{1}{5} \exp(t^2)$ results in bounds by subgaussian tails decreasing like $\exp(-C t^2)$, for some constant $C$.

We next introduce the tail inequalities that use a standard chaining argument to control uniform convergence over $\pi \in \Pi$. Let $n$ denote a generic dataset size (we will later on apply the results with $n=NT$.) The data are $(X_{1:n}, A_{1:n}, Y_{1:n})$ and $f_i(\pi)$ is a function of $(X_i, A_i, Y_i)$. Define the function class	$\mathcal{F}(X_{1:n}, A_{1:n},Y_{1:n}) = \{ (f_i(\pi), \dots , f_{n}(\pi)) \colon \pi \in \Pi  \}. $

For this section, we consider maximal inequalities for the function classes for the enveloped policy class $\mathcal F$. Let $\epsilon_i \in \{-1,+1\}$, be iid Rademacher variables (symmetric Bernoulli random variables with value $-1,+1$ with probability $\frac 12$), distributed independently of all else. We use the following application of chaining with a bounded envelope function, due to  \cite[Eqn. 7.3]{pollard1990empirical}. (Using different measures of functional complexity for multi-class predictors, such as Natarajan dimension, simply changes the constants in the final bound.)

\begin{theoremlit}{A}[Uniform convergence of policy function $\pi$ over envelope class $\mathcal F $. ]\label{lemma-policy-uc}
	Let \mbox{$f(\pi)\leq \norm{F}_2 \leq C$} be a bound on the envelope function for $f \in \mathcal F$. 
	Then for $n$ large enough, where $\vc$ is the VC-dimension,  
	\begin{equation} \sup_{ f \in \mathcal F } \abs{ \frac{1}{n}\sum_{i=1}^n (f_i(\pi) - \E[f(\pi)] ) } \leq  \nicefrac{9}{2} C  \sqrt{\frac{ \vc \log(\nicefrac{5}{\delta}) }{{n} } }
	\end{equation}
\end{theoremlit}

The next variant is a modification of Thm. \ref{lemma-policy-uc} which only uses only moment bounds for the envelope function at the expense of weaker controls of the tails of the supremum process.
\begin{theoremlit}{B}[Uniform convergence with $L_p$ norm of envelope function.]\label{lemma-policy-uc-envelope} 
For an absolute constant $C_{\vc}$ which depends only on $\vc$, where $F_n$ is the envelope for $f\in \mathcal F$ and $\vc$ is the VC dimension,
$ \textstyle
\E[\sup_{ f \in \mathcal F } \abs{ \frac{1}{n}\sum_{i=1}^n (f_i(\pi) - \E[f(\pi)] ) } ] \leq 
C_{\vc}' \sqrt{\vc-1}  \E[ \abs{F_n} ]$.
\end{theoremlit}

	We also state a standard lemma used for sample splitting, as appears in \cite{chernozhukov2018double}, without proof.
\begin{lemma}[Conditional convergence implies unconditional]
	Let $\{ X_m\}$ and $\{Y_m\}$ be sequences of random vectors. 
	\begin{itemize}
		\item If for $\epsilon_m \to 0$, $\pr(\norm{X_m} > \epsilon_m \mid Y_m) \to_p 0,$ then $\pr(\norm{X_m} > \epsilon_m) \to 0.$ This occurs if $\E[ \norm{X_m}^q/\epsilon_m^q \mid Y_m ] \to_p 0$ for some $q \geq 1$ by Markov's inequality. 
		\item Let $\{A_m\}$ be a sequence of positive constants. If $\norm{X_m} = O_p(A_m)$ conditional on $Y_m$, that for any $\ell_m \to \infty$, $\pr(\norm{X_m} > \ell_m A_m \mid Y_m)\to_p 0,$ then $\norm{X_m} = O_p(A_m)$ unconditionally, namely, that for any $\ell_m \to \infty$, then $\pr(\norm{X_m} > \ell_m A_m)\to 0$. 
	\end{itemize}
\end{lemma}

\begin{proof}[Proof of Thm.~\ref{lemma-policy-uc}]
	We first bound the deviations uniformly over the policy class and introduce the following empirical processes, $$M = \sup_{ f \in \mathcal F } \abs{ \sum_{i=1}^n (f_i(\pi) - \E[f(\pi)] ) }, \quad
	L = \sup_{ f \in \mathcal F }  \abs{ \sum_{i=1}^n \epsilon_i f_i(\pi)  }.$$ 
	By a standard symmetrization argument, applying Jensen's inequality for the convex function $\Phi$ of the symmetrized process (e.g. Theorem 2.2 of \cite{pollard1990empirical}), we may bound the Orlicz norm of the maxima of the empirical process by the symmetrized process, conditional on the observed data: 
	$ \E[ \Phi(M)] \leq  \E[ \Phi(2L)] .  $
	Taking Orlicz norms with $\Phi(t) = \frac{1}{5} \exp(t^2)$, we apply a tail inequality on the Orlicz norm of the symmetrized process $\Phi \left( 2 L  \right)$, under the assumption of bounded outcomes. Applying Dudley's inequality to the symmetrized empirical process $L$, (e.g. Theorem 3.5 of \cite{pollard1990empirical}), we have that 
	\begin{equation}\label{duddleyinteq} \E_\epsilon\bracks{ \exp(L^2/J^2) \mid \mathcal D }\leq5~~\text{for}~~ J=9\magd{F}_2\int_0^1\sqrt{\log(D(\magd{F}_2\zeta, \mathcal F(X_{1:n}) ))}d\zeta. \end{equation}
	
	By Markov's inequality, we have that
	$ \mathbb{P}\left( \frac{1}{n} L > t \right) \leq 5 \exp(-t^2n^2/ \norm{L}^2_2 ) = 5 \exp( -t^2 n /J^2 C^2  ), $
	so that therefore, bounding the Dudley entropy integral by the VC dimension via \cite[eqn. 7.8]{pollard1990empirical},
	$$ \frac{1}{n}M \leq \frac{  \nicefrac{9}{2} C \sqrt{\vc\log(\nicefrac{5}{\delta})} }{\sqrt{n} } . $$
\end{proof}

\begin{proof}[Proof of Thm.~\ref{lemma-policy-uc-envelope}]
	By standard results on covering numbers and VC dimension, e.g. \cite{van1996weak,wainwright2019high}, for a VC-class of functions with measurable envelope function $F$ and $p\geq 1$, for any probability measure $Q$ with $\norm{F}_{Q, p}>0$, $N(\epsilon \norm{F}_{Q,p}, \mathcal F, L_p(Q)) \leq A(\vc) (\frac{1}{\epsilon})^{p(\vc -1)}$, where $C_{\vc})$ is an absolute constant that depends only on the vc-dimension. By \cite[eqn. 7.8]{pollard1990empirical}, 
\begin{align*}\textstyle
	\E[ \sup_\pi  \abs{ \frac 1n \sum_{i=1}^n (f_i(\pi) - \E[f(\pi)] ) }^p  ]
	&
	\leq 
	\left(18 C_p \int_0^1 \sqrt{\log\left(C_{\vc} \left(\nicefrac{1}{\epsilon}\right)^{p(\vc -1)} \right) } dx  \right) \E[ \abs{F_n}^p ] 
	\\
	&
	\leq 
	C_{\vc}' \sqrt{\vc-1}  \E[ \abs{F_n}^p ]
	\end{align*}
The claim follows by taking $p=2$. 
\end{proof}

	\paragraph{Preliminaries }

	A key step of the analysis is leveraging an additive decomposition of the regret for finite horizons. For example, this appears in \cite{jin2018q}; we simply include the full statement for completeness and verify for our setting. Recall that we denote $\margP$ as the transition matrices and estimated transition matrices corresponding to the marginal MDP (e.g. transitions between system states). In this section, for brevity we let $\hatmargP$ denote the empirical counterpart of $\margP$, e.g. estimating \cref{eqn-batch-policy-value} via some estimation strategy (IPW weighting, doubly robust, or plug-in estimation; typically we focus on the doubly-robust estimator). Similarly, $\widehat{M}$ denotes the empirical MDP model with $\hatmargP$. We introduce notation for indexing into entries after evaluating the transition operator, $$(\margP V)(s,\pi) = \E_{s' \sim \margP(\cdot\mid s,\pi)} V(s') = \E[ \margR(s,\pi) ]+ \sum_{(s',y) \mid s}    P(Y=y\mid \pi) (V(s'))$$
	We also introduce notation for indexing the difference between applying the true transition operator and the empirical estimate thereof, for any \textit{generic} $\nS$-vector v and policy $\pi$: 
	$$((\margP -\hatmargP)v)(s,\pi) =  \sum_{(s',y) \mid s}  \left( P(Y=y\mid \pi)  - \hat{P}(Y=y\mid \pi) \right)v(s') .$$

	\begin{lemma}[Additive decomposition of finite-horizon value iteration]\label{lemma-value-func-additive-decomp} 
		For any policies $\tilde\pi,\pi$ and any $(s,t') \in \Ss \times [T]$, 
		\begin{align*}&\tildeV_{t'}^{\tilde\pi} (s) - \htpiV_{t'}(s)
			\\ &= \E_{\hat M, \tilde\pi} \left[ 
			\sum\nolimits_{t=t'}^{\horizon} \E[ \margR(s,\pi_t) ] - \hat \E[ \margR(s,\pi_t) ] + \sum\nolimits_{(s',y) \mid s}  \left( P(Y=y\mid \pi_t) - \hat{\pr}(Y=y\mid \pi_t)\right)\tildeV_{t+1}^{\tpi}(s') \mid s_{t} = s \right] 
		\end{align*} 
	\end{lemma}

	\subsection{Sample complexity analysis}

	We first provide a generalization bound in an \textit{oracle nuisance} case where we use the true conditional expectations $\mu, e$ rather than estimated counterparts $\hat \mu, \hat e$. 
	\begin{align}\Gamma_t^{i,*}(y\mid a) &\defeq
		\frac{\mathbb{I}[Y_{t}^i=y]-{\mu}(y \mid A_t^{i}, X_t^{i})}{{e}\left(A_t^{i}\mid X_t^{i}\right)}
		\mathbb{I}[A_t^i=a]
		+{{\mu}} (y \mid a , X_t^i) \label{eqn-oracle-nuisance} 
		\\
		\hpypi
		& 
		\textstyle
		\defeq \frac{1}{N\horizon} \sum_{t=1}^\horizon 	\sum_{i=1}^N  \sum_{a \in \Aa} \pi(a \mid X_t^i) \hat {\Gamma}_t^{i,*}(y\mid a)     \nonumber
	\end{align}

	\begin{theorem}[Sample complexity
		and rate double-robustness for oracle estimator ]\label{thm-policy-learning-oraclesample-complexity} 
		Suppose ${e(a\mid x)\leq \nu^{-1}}$ uniformly over $a,x$ (overlap).
		
		Then w.p. $\geq 1-\delta$, for $ \tildeV^{\hat\pi^*}_0$ optimized via \Cref{alg-backwards-recursion} with oracle nuisance estimator \cref{eqn-oracle-nuisance},
		$$\tildeV^{\pi^*}_0 - \tildeV^{\hat\pi^*_{\op{oracle}}}_0 	\leq \frac{   \nu \vc \Rmax( \horizon  + \nicefrac{1}{2}\horizon^2) \nicefrac{9}{2}  \sqrt{\log(\nicefrac{5 \horizon \vert \Yy \vert }{\delta})} }{\sqrt{N \horizon} } $$
	\end{theorem}

\begin{proof}[Proof of \Cref{thm-policy-learning-oraclesample-complexity}]
	
	In the analysis, we leverage single-timestep uniform convergence arguments. By \Cref{asn-X-exogeneous-context}, $X$ is drawn exogenously/independently of all else, so the $X$ data is iid. We model the dataset as drawn from multiple behavior policies, so that at timestep $t$, $S \sim \rho_t^{\pi_b}(S), X\sim f_t, A\sim \pi_b(s,x)$, and $Y\mid A,X$ is drawn from the contextual response model. Therefore the data tuples $(S_{i,t}, X_{i,t},A_{i,t},  Y_{i,t})$ are viewed as independent draws from this process. Since $Y\mid A,X \indep S_t$ and therefore we only need to control for $X_t$ such that $e(X_t)$ is not a function of $S$, the functions we use in our estimator, defined with respect only to the $(X_{i,t}, A_{i,t}, Y_{i,t})$ data, admit analysis via iid/single-stage empirical process techniques. 
	
	We define the following function classes conditional on all the data, $(X_{1:NT},A_{1:NT}, Y_{1:NT})$. 
	For $\pi$, we consider a shifted function class with an envelope function: let ${f_{it}(\pi) =  \pi(A^i_t\mid X^i_t) \mathbb{I}[A^i_t = a]  \mathbb{I}[Y^i_t=y]}$ where
	$$\mathcal{F}(X_{1:NT}, A_{1:NT},Y_{1:NT}) = \{ (f_t^i(\pi), \dots , f^{N}_{\horizon}(\pi)) \colon \pi \in \Pi  \}. $$
	
	\underline{Step 1:} Error decomposition:

	We decompose the error. By optimality of $\hat{\tpi}^*$, $\htPiV{\tpi^*}_0-\htPiV{\hat{\tpi}^*}_0 \leq 0$ and the triangle inequality: 
	\begin{align}
		\tildeV_0^{\tpi^*} - \tildeV_0^{\widehat{\tpi}^*} & = \tildeV_0^{\pi^*} - \htPiV{\tpi^*}_0
		+\htPiV{\tpi^*}_0-\htPiV{\hat{\tpi}^*}_0 + \htPiV{\hat{\tpi}^*}_0 
		- \tildeV_0^{\hat{\tpi}^*}  
		\nonumber	
		\\
		&\leq \abs{\tildeV_0^{\tpi^*} - \htPiV{\tpi^*}_0} + \abs{\htPiV{\hat{\tpi}^*}_0 
			- \tildeV_0^{\hat{\tpi}^*}} \label{eqn-decomposition} 
	\end{align}

	\underline{Step 2:} Uniform convergence over $\pi$ and $\tildeV^\pi$:

We apply the additive decomposition of \Cref{lemma-value-func-additive-decomp} and obtain a uniform bound on 
	$\sup_{\pi \in \Pi}  \tildeV_{0}^{\tilde\pi} (s) - \htpiV_{0}(s)   $, which we apply twice to the terms of \Cref{eqn-decomposition}. 
	
	For $\pi$, we consider a shifted function class with an envelope function: define 
	\begin{align*}f_i(\pi) &=  \pi(A_i\mid X_i) \mathbb{I}[A_i = a]  \mathbb{I}[Y_i=y] e^{-1}(a\mid X_i )\\
		\mathcal{F}(X_{1:n}, A_{1:n},Y_{1:n}) &= \{ (f_1(\pi), \dots , f_{n}(\pi)) \colon \pi \in \Pi  \}. 
	\end{align*}

	By \Cref{lemma-value-func-additive-decomp},	for any policy $\pi$, consider the additive decomposition:	$$ \sup_{\pi \in \Pi}  \tildeV_{0}^{\tilde\pi} (s) - \htpiV_{0}(s)  =\sup_{\tilde\pi \in \Pi}  \E_{\hat M, \tilde\pi} \left[ 
	\sum_{t=0}^{\horizon}	(\margR(s,\tpi_t) -  \hatmargR(s,\tpi_t)  )  + (\margP_{t} - \hatmargP_{t} )( \tildeV_{t+1}^{\tpi})(s_{t}, \tpi_t) \mid s_{t} = s \right] $$
	Note that the dependence on the state distribution is not relevant (that is, the expectation under $(\hat M, \hat \pi)$ since the estimator uses the same data at every state $s$, and hence it suffices to consider uniformity over $\Yy$ and equivalently evaluate the supremum over corresponding transitions.
	\begin{align}\label{eqn-samplecomplexity-decomposition}
		&  	\sup_{(\tpi, \tpi') \in \Pi \times \Pi} 
		\sum_{t=0}^{\horizon} \sup_{y \in \Yy} 
		\left\{
		(\margR(s,\tpi_t) - \hatmargR(s,\tpi_t) ) +
		(
		P(Y=y\mid \pi_t)  - \hat P(Y=y\mid \pi_t) 
		)( \tildeV_{t+1}^{\tpi'_{t+1:T}}))(s_{t}, \pi_{t}) \right\} \nonumber \\
		&\leq
		\sum_{t=0}^{\horizon}
		\sup_{(\tpi_t, \tpi^{(t)}_{t+1:T}) \in \Pi_t \times \Pi_{t+1:T}} \sup_{y \in \Yy} 
		\left\{
		(\margR(s,\tpi_t) - \hatmargR(s,\tpi_t) ) +
		(
		P(Y=y\mid \pi_t)  - \hat P(Y=y\mid \pi_t) 
		)( \tildeV_{t+1}^{\tpi^{(t)}_{t+1:T}}))(s_{t}, \pi_{t}) 
		\right\}\\
		&		\leq
		\sum_{t=0}^{\horizon}
		\sup_{y \in \Yy} 
		\left\{
		\sup_{\tpi_t \in \Pi_t } 		\abs{(\margR(s,\tpi_t) - \hatmargR(s,\tpi_t) )} +
		\sup_{\tpi_t \in \Pi_t } 		\abs{
			 (T-t-1)\cdot \left(	P(Y=y\mid \pi_t) - \hat P(Y=y\mid \pi_t) \right)
		}
		\right\}
	\end{align}
	Since forward transitions marginalize to $\tildeV_{t+1}(s')$, in the last line we observed that taking the supremum over $\pi_{t+1:T}$ can only enlarge the fixed envelope function, $\norm{F}_2$, but does not actually affect the empirical process analysis. Under assumption of a product set of policies across states, uniform convergence under $\tpi_t \in \Pi_t$ also establishes uniform convergence for state-dependent policies. 
	
	\underline{Step 3}: applying the concentration inequalities. 
	We bound each of the above terms by Thm.~\ref{lemma-policy-uc}, applying a high probability bound with $\delta' = \frac{\delta}{\horizon}$, and finally take a union bound over $\Yy$ and each summand, in order to obtain the following bound which holds with probability $> 1-\delta$,
	\begin{align*}  \sup_{\tilde\pi \in \Pi} \tildeV_{0}^{\tilde\pi} (s) - \htpiV_{0}(s)  &\leq \sum_{t=0}^H \frac{  \nu v \Rmax (q + (\horizon - t -1)) \nicefrac{9}{2}\sqrt{\log(\nicefrac{5\horizon\vert \Yy \vert }{\delta})} }{\sqrt{N \horizon} }  \\
		& = \frac{   \nu v \Rmax( \horizon  + \nicefrac{1}{2}\horizon^2) \nicefrac{9}{2}  \sqrt{\log(\nicefrac{5 \horizon \vert \Yy \vert }{\delta})} }{\sqrt{N \horizon} } 
	\end{align*} 
	
\end{proof}
	
	\begin{proof}[Proof of \Cref{thm-policy-learning-sample-complexity-uniform-convergence}]
		
		We decompose the regret achieved by the \textit{feasible} estimator as: 
		
		$$\tildeV^{\pi^*}_0 - \tildeV^{\hat\pi^*_{\op{feasible}}}_0  = 
		\tildeV^{\pi^*}_0 - 
		\tildeV^{\hat\pi^*_{\op{oracle}}}_0 +  \tildeV^{\hat\pi^*_{\op{oracle}}}_0 -
		\tildeV^{\hat\pi^*_{\op{feasible}}}_0.  $$
		
		\Cref{thm-policy-learning-oraclesample-complexity} established the bound on $\tildeV^{\pi^*}_0 - 
		\tildeV^{\hat\pi^*_{\op{oracle}}}_0$. We now show that $\tildeV^{\hat\pi^*_{\op{oracle}}}_0 -
		\tildeV^{\hat\pi^*_{\op{feasible}}}_0 = o_p(n^{-\frac 12})$ under the rate assumptions on $\hat e, \hat \mu$. 
		
		Verifying rate double-robustness is standard given arguments in the single-timestep literature. The key step is, as in the proof of \Cref{thm-policy-learning-oraclesample-complexity}, applying the additive error decomposition of \Cref{lemma-value-func-additive-decomp}. 
		
		Then standard single-timestep analysis for doubly-robust policy optimization yields the result \cite{wager17,zhou2018offline}. We simply apply our uniform convergence bounds and state the decomposition for completeness.

		Let $ \hat \pr(Y=y\mid a)$ denote the estimator for the policy value. Write 
		$\hat \pr^{(1)}(Y = y \mid a) $ for the estimator evaluated on the first fold using out-of-fold nuisances, and vice-versa.

		\begin{align*}\hat\Gamma_t^i(y\mid a) &\defeq
			\frac{\mathbb{I}[Y_{t}^i=y]-\hat{\mu}^{-k(i,t)}(y \mid A_t^{i}, X_t^{i})}{\hat{e}_{t}^{-k(i,t)}\left(A_t^{i} \mid X_t^{i}\right)}
			\mathbb{I}[A_t^i=a]
			+\hat{{\mu}}^{-k(i,t)}(y \mid a , X_t^i)\\
			\hat\tau(a,y)
			&=\frac 1n	\sum_{i=1}^N  \sum_{a \in \Aa} \pi(a , X_t^i) \hat {\Gamma}_t^i(y\mid a)     
		\end{align*} 
		
		We denote partial sums of the individual contribution to the estimator $\hat\Gamma_t^i$, as $\hat \Gamma_{\mathcal I_1}$, 
		in order to denote the sum over the (sample-split) dataset evaluated with the estimated nuisances, and $\Gamma_{\mathcal I_1} $ denote evaluation of the corresponding estimator summands over the oracle nuisance functions. 
		
		Due to sample splitting, conditional on a fold $\mathcal I_1$, $\hat\mu^{(1)}$ can be treated as deterministic. 
		The decomposition used in the proof of \Cref{thm-policy-learning-sample-complexity-uniform-convergence}, \cref{eqn-samplecomplexity-decomposition} allows us to study regret of the oracle vs. nuisance estimators relative to the {true} value function $\tildeV$. We establish uniformity of the error from using estimated nuisances: 
		\begin{equation}
			\sup_{\pi_{t:T}} 
			\abs{  \sum_{a\in \Aa}
				\sum_{i=1}^n  \pi_t(a\mid X_i) (\hat\Gamma^i_{t}(y\mid a) - \Gamma^i_{t}(y\mid a) ) (R(s,a,s')+  V^{\pi_{t+1:T}}_{t+1}(s'))
			} = o_p(n^{-\frac 12})
			\label{eqn-samplecomplexity-ratedr-feasible} 
		\end{equation}
		
		We decompose the terms as follows, restricting attention to one action, by adding and subtracting $ \frac{(\mathbb{I}[Y_{t}^i=y]-{\mu}^{-k(i)}(a, X_t^{i}))}{\hat e^{-k(i)} (a, X^i_t)}$. To clear up the display we suppress arguments depending on $X_i$ and others where self-evident. 
		\begin{align}
			& 
			\E_{N,T} [\pi(a) (	\hat\Gamma_{\mathcal I_1}^i - \Gamma_{\mathcal{I}_1}^i )]
			=
			\nonumber\\
			&
			\E_{N,T} \left[\pi(a)
			\left(
			\left( \hat{{\mu}}^{-k(i,t)}- {{\mu}}^{-k(i,t)}	\right) + 
			\left(
			\frac{\mathbb{I}[Y_t=y]-\hat{\mu}^{-k(i,t)}}{\hat{e}_{t}^{-k(i,t)}}-
			\frac{\mathbb{I}[Y_t=y]-{\mu}^{-k(i,t)}}{{e}_{t}^{-k(i,t)}}
			\right)
			\mathbb{I}[A_t=a]
			\right)
			\right] \nonumber\\
			& = 
			\E_{N,T} \left[ \pi(a)
			\left( \hat{{\mu}}^{-k(i,t)}- {{\mu}}^{-k(i,t)}	\right) 
			\left(1 - \frac{ \mathbb{I}[A_t=a]}{{e}_{t}^{-k(i,t)}}
			\right)(R+  V^{\pi_{t+1:T}}_{t+1}(s'))
			\right]
			\label{eqn: rateDR 1} \\
			& 
			+\E_{N,T} \left[  
			\pi(a)
			\left( {{\mu}}^{-k(i,t)}- \hat{{\mu}}^{-k(i,t)}	\right) 
			\left(\frac{ \mathbb{I}[A_t=a]}{{e}_{t}^{-k(i,t)}} - \frac{ \mathbb{I}[A_t=a]}{\hat{e}_{t}^{-k(i,t)}}
			\right) (R+  V^{\pi_{t+1:T}}_{t+1}(s'))
			\right]
			\label{eqn: rateDR 2}
			\\
			&  
			+\E_{N,T} \left[
			\pi(a)
			(\mathbb{I}[Y_t=y]-{\mu}^{-k(i,t)})
			\left(
			\frac{1}{\hat{e}_{t}^{-k(i,t)}}-
			\frac{1}{{e}_{t}^{-k(i,t)}}
			\right)(R+  V^{\pi_{t+1:T}}_{t+1}(s'))
			\right]
			\label{eqn: rateDR 3}
		\end{align}
		
		Decomposing each of the above terms into foldwise terms, sample splitting implies that conditioning on the other folds implies that $\mu_1$ is a deterministic function.  
		
		We apply our \Cref{thm-policy-learning-sample-complexity-uniform-convergence} to establish $o_p(n^{-\frac 12})$ rates on the relevant uniform convergence terms of \cref{eqn: rateDR 1,eqn: rateDR 2}. 
		
		The term of \cref{eqn: rateDR 1} evaluates to $0$ by iterating expectations, using independent errors property from cross-fitting, and well-specification of $e$ which implies $
		0=\E_{N,T} \left[
		\left(1 - \nicefrac{ \mathbb{I}[A_t=a]}{{e}_{t}^{-k(i,t)}}\right) \mid X
		\right]$.
		
		The term of \cref{eqn: rateDR 2} is bounded by Cauchy-Schwarz since $\pi \leq 1$ and by assumption on the sum of rates in \Cref{thm-policy-learning-sample-complexity-uniform-convergence}: \cref{eqn-dr-muunbiased-2} 
		\begin{align*}
			\sup_{\pi} (\ref{eqn: rateDR 2})
			&\leq \E_{N,T} 
			\left[
			\abs{\mu^{-k(i,t)} - \hat\mu^{-k(i,t)} }		
			\abs{\frac{ \mathbb{I}[A_t=a]}{{e}_{t}^{-k(i,t)}} - \frac{ \mathbb{I}[A_t=a]}{\hat{e}_{t}^{-k(i,t)}}
			} \right] \\
			&\leq 
			\E_{N,T} 
			[(\mu^{-k(i,t)} - \hat\mu^{-k(i,t)} )^2	]^{\frac 12}	
			\E_{N,T} 
			\left[
			\left(\frac{ \mathbb{I}[A_t=a]}{{e}_{t}^{-k(i,t)}} - \frac{ \mathbb{I}[A_t=a]}{\hat{e}_{t}^{-k(i,t)}}
			\right)^2 
			\right]^{\frac 12}	\\
			&=  o_p(n^{-\frac 12})
		\end{align*} 
		
		For the term of \cref{eqn: rateDR 3}, we apply the bound of Thm.~\ref{lemma-policy-uc-envelope} which surfaces the dependence of the maximal inequality on the behavior of the envelope function, allowing us to leverage consistency assumptions of \Cref{thm-policy-learning-sample-complexity-uniform-convergence} to verify that the term is $o_p(n^{-\frac 12})$. 
		\begin{align*}
			\E[ \sup_\pi (\ref{eqn: rateDR 3}) ]& 	\leq \E_{N,T}[ \pi(a) (\mathbb{I}[Y_t=y]-\mu^{-k(i,t)}) (R +V_{t+1}^{\pi_{t+1:T}}(s'))   ] \E_{N,T} 
			\left[	\abs{
				({\hat{e}_{t}^{-k(i,t)}})^{-1}-
				({{e}_{t}^{-k(i,t)}})^{-1}
			}\right]  \\
			& \leq
			\E_{N,T}[ \pi(a) (\mathbb{I}[Y_t=y]-\mu^{-k(i,t)}) (R +V_{t+1}^{\pi_{t+1:T}}(s'))   ]  \times \nu { \E_{N,T} 
				\left[	
				\abs{{\hat{e}_{t}^{-k(i,t)}}-
					{{e}_{t}^{-k(i,t)}}
				}\right] } \\
			&  = o_p(n^{-\frac 12})
		\end{align*}
		\cref{eqn-samplecomplexity-ratedr-feasible} follows from the above and applying Markov's inequality for the bound for \cref{eqn: rateDR 3}. 
	\end{proof}
	
	\begin{proof}[Proof of \cref{lemma-value-func-additive-decomp}]
	
	We follow an induction argument. Note that since $\tildeV_{T+1}=0$, the base case, $t=T$, satisfies that 
	$$\tildeQ_{T}^{\tpi_T}(s,\pi_T) - \htQ_{T}^{\tpi_T}(s,\pi_T) =  \E[ \margR(s,\pi_T) ] - \hat \E[ \margR(s,\pi_T) ].$$
	
	Now suppose the inductive hypothesis holds for $t+1$ and consider the case of $t$. 
	\begin{align*} \tildeQ_t^\pi(s,\tpi_t)-\htQ_t^\pi(s,\tpi_t) &
		=\margR(s,\tpi_t) - \hatmargR(s,\tpi_t)  + (\margP \tildeV_{t+1}^{\tpi})(s,\tpi_{t}) - (\hatmargP \htV_{t+1}^{\tpi})(s,\tpi_{t})\\
		& =
		\margR(s,\tpi_t) - \hatmargR(s,\tpi_t)  + ((\margP -\hatmargP)\tildeV_{t+1}^{\tpi})(s,\tpi_t) - (\hatmargP (\tildeV_{t+1}^{\tpi}-\htpiV_{t+1}))(s,\tpi_t)
	\end{align*}
	by a standard performance difference lemma. Since actions in the policy space are themselves policies, the previous analysis for $Q$ functions applies also to $V$ functions, 
	\begin{align*}
		\tildeV^{\tpi}_t (s) - \htpiV_t (s) &= \E_\pi \left[
		\margR(s,\tpi_t) - \hatmargR(s,\tpi_t)  + ((\margP -\hatmargP)\tildeV_{t+1}^{\tpi})(s_t,\tpi_t) \mid s_t =s 
		\right]
		\\& \qquad 
		+
		\E_{\hat M, \tpi_t} \left[
		\tildeV^{\tpi}_{t+1}(s_{t+1}) - \htPiV{\tpi}_{t+1}(s_{t+1}) \mid s_t = s
		\right] \\
		&  = \E_{\hat M, \tilde\pi} \left[ 
		\sum_{t=t'}^T 	(\margR(s,\tpi_t) - \hatmargR(s,\tpi_t) )  + (\margP_{t} - \hatmargP_{t} )( \tildeV_{t+1}^{\tpi})(s_{t}, \tpi_t) \mid s_{t}= s \right] 
	\end{align*}
	where we apply the inductive hypothesis in the first line. \qedhere
\end{proof}

\subsection{Verifying nuisance function rates from pooled episodes}\label{apx-sec-nuisancerates}

\paragraph{Summary of argument}
We simply verify that the nuisances can be learned from the pooled episodes at the relevant rates. Note that we learn $\mu(a\mid x)$ from data that can be viewed as the collection of $\{ (X_t^i, A_t^i, Y_t^i)\}^n_{0:\horizon},$
and $e(a\mid x)$ from $\{ (X_t^i, A_t^i)\}^{n}_{0:\horizon}$.
Throughout the paper, we assume that the behavior policy is stationary (i.e. we do not handle dependent data from history-adapted policies, such as from learning policies, though doing so is a straightforward extension via standard mixing arguments). Hence, the sequence of states is strongly stationary. Nonetheless, dependency across timesteps within an episode is a consequence of $S_t,X_t$ adapted policies (although there is no dependence in the noise of the contextual response across timesteps). 

Our argument proceeds as follows. First, we invoke results establishing the mixing properties of the infinite-horizon embedding of episodic finite-horizon MDPs. This general result of \cite{bojun2020steady} provides a construction (with a modification to ensure aperioridicity that preserves other properties of the chain) ensures that such an embedding has a steady-state distribution. Therefore, we model nuisance estimation as if we estimate the nuisance functions from a single infinite-horizon trajectory, obtained via the construction of \cite{bojun2020steady} by simply sequentially concatenating the episodes (and the perturbations for aperiodicity). In this infinite-horizon embedding, the stationary distribution is given by the (finite-horizon) limiting state-action frequencies, so that the sequence is $\beta-$mixing. (Clearly, such an argument generalizes to the case of history-adapted policies with a corresponding dependence on mixing rate of the adapted policy). We invoke results on learning from $\beta$-mixing sequences, e.g. \cite{farahmand2012regularized}, which in particular applies the blocking empirical process argument with a more refined peeling empirical process argument. 

\subsubsection{Preliminaries} 

We quote a result on learning from $\beta$-mixing data of \cite{farahmand2012regularized}. The assumptions and result are stated for a generic conditional expectation on a generic process $\{ X_{t}, Y_{t} \}_{0:T}^{1:N}$. (We of course instantiate the result for the processes $\{X_{t}^i, A_t^i, Y_{t}^i\}_{0:T}^{1:N}, \{X_{t}^i, A_t^i\}_{0:T}^{1:N}$). The estimator of interest is regularized least-squares regression $\hat m_n(x)= \op{clip}_{-L,L}(\tilde m(x))$ where the final estimates are clipped within a bounded range, and $J$ is a regularization functional. 
\begin{equation}\tilde{m}_{n}=\underset{f \in \mathcal{F}}{\operatorname{argmin}}\left\{\frac{1}{n} \sum_{i=1}^{n}\left|f\left(X_{i}\right)-Y_{i}\right|^{2}+\lambda_{n} J^{2}(f)\right\}
	\label{eqn-nuisance-regls}
	\end{equation}
\begin{assumption}[Exponential Mixing]\label{asn-dependent-exponentialmixing}
The process $\left(\left(X_{t}, Y_{t}\right)\right)_{t=1,2, \ldots}$ is an $\mathcal{X} \times \mathbb{R}$-valued, stationary, exponentially $\beta$-mixing stochastic process. In particular, the $\beta$-mixing coefficients satisfy $\beta_{k} \leq \bar{\beta}_{0} \exp \left(-\bar{\beta}_{1} k\right)$, where $\bar{\beta}_{0} \geq 0$ and $\bar{\beta}_{1}>0$
\end{assumption}
\begin{assumption}[Capacity]\label{asn-dependent-capacity}
	There exist $C>0$ and $0 \leq \alpha<1$ such that for any $u, R>0$ and all $x_{1}, \ldots, x_{n} \in \mathcal{X}$
	$$
	\log \mathcal{N}_{2}\left(u, \mathcal{B}_{R}, x_{1: n}\right) \leq C\left(\frac{R}{u}\right)^{2 \alpha}
	$$
\end{assumption}
\begin{assumption}[Boundedness]\label{asn-dependent-boundedness}
There exists $0<L<\infty$ such that the common distribution of $Y_{t}$ is such that $\left|Y_{t}\right| \leq L$ almost surely.
\end{assumption}
\begin{assumption}[Realizability]\label{asn-dependent-realizability}
The regression function $m(x)=\mathbb{E}\left[Y_{1} \mid X_{1}=x\right]$ belongs to the function space $\mathcal{F}$.
\end{assumption}

We verify we may learn the nuisance functions in the following statement, which is a corollary of the construction of \cite[Thm. 2]{bojun2020steady} and the convergence rate of \cite[Thm. 5]{farahmand2012regularized}. 
\begin{corollary*}[Nuisance functions]
Suppose nuisance function estimator $\hat e(a\mid x)$ is learned from pooled-episode data $\{ (X_t^i, A_t^i, Y_t^i)\}^n_{0:\horizon},$  
and $\mu(y\mid a, x)$ from $\{ (X_t^i, A_t^i)\}^{n}_{0:\horizon}$. Let A2-A4 hold%
, then there exists $c_1, c_2>0$ such that for large enough $n$, for any fixed $\delta \in (0,1)$, with probability $\geq 1-\delta$,
$
\E[(m(X) - \hat m(X))^2] \leq c_{1}\left[J^{2}(m)\right]^{\frac{\alpha}{1+\alpha}} n^{-\frac{1}{1+\alpha}}\left[\frac{\log \left(n \vee c_{2} / \delta\right)}{\bar{\beta}_{1}}\right]^{3}
$.
\end{corollary*}
\begin{proof}[Proof of \Cref{cor-nuisance-rates}.]

	We first consider the infinite-horizon embedding, $\mathcal M^L$, of the finite-horizon MDP. To keep the presentation self-contained, we describe the construction of \cite{bojun2020steady}. To allow for time-inhomogeneous (but state-stationary) policies, augment the state space with a time state $T_t$ so that the infinite-horizon MDP's state space is $(T_t,S_t, X_t)$. We embed the finite-horizon MDP by advancing time states in the natural sense, and transitioning from end-of-episodes to initial states,  \begin{align*}P(t', s_{t+1}, x_{t+1}\mid t,  s_t, x_t, a_t ) &= \mathbb{I}[t'=t+1]
		P(s_{t+1}, x_{t+1}\mid t,  s_t, x_t, a_t ) \\
		P({k\horizon+1}, s_0, x_{k\horizon+1}\mid kH,  s_{k\horizon}, x_{k\horizon}, a_{k\horizon} ) &= P(s_0 ), \forall k \in \mathbb{N}
	\end{align*}
	Such a process satisfies the definition of an episodic learning process, e.g. definition 1 of \cite{bojun2020steady}. Now, to ensure ergodicity due to periodicity of the episodic/finite-horizon structure, \cite{bojun2020steady} establishes that a simple $\epsilon-$perturbation recovers aperiodicity by introducing a perturbed MDP, $\mathcal{M}^{+}$ which simply introduces an auxiliary null state $s_{null}$.  With some $\epsilon$ probability, transitions from terminal state (e.g. $T_t \text{ mod }\horizon = 0$) transit to $s_{null}$ before transiting to the initial state and beginning another episode. Theorems 2 and 3 verify ergodicity (evident under aperiodicity) and that the $Q$ functions are identical between $\mathcal{M}^L$ and $\mathcal{M}^+$. Properties of the construction are clear under aperiodicity, see, e.g. \citep{meyn2012markov}.
	
	Clearly, the stationary distribution of the chain is $$p^{\infty,+}_{\pi}(t, s_t, x_t, a_t) = \frac{1}{\horizon} f_t(x_t)d^{\pi}(s_t) \pi_t(a_t\mid s_t,x_t),$$ where $d^\pi(s_t)$ is the time-t marginal state occupancy distribution under $\pi$ in the original finite-horizon MDP, $\mathcal M$. Convergence to stationarity distribution of $\mathcal{M}^L$ (and hence $\mathcal{M}^+$) is convergence of empirical state-occupancy distributions $d^{\pi}(s_t)$, which converge geometrically.

\end{proof}

\subsection{Proofs of structural analysis for dynamic/capacitated pricing}
We include the full statement of theorem including the general sufficient condition for $t< T-2$ that is less interpretable.

\begin{assumption}[Reward ordering]
	
	$R(1)>R(0)$, without loss of generality. \label{asn-structuralanalysis-rewardorder}
\end{assumption}
\begin{assumption}[Discrete concavity of value function]
	$V_t$ is a discrete concave function in $s$, for all $t$.  \label{asn-structuralanalysis-concavity} 
\end{assumption}
\begin{assumption}[Unimodal expected reward ] \label{asn-structuralanalysis-unimodality}

	$\tildeR(s,\pi)$ is decreasing for suboptimal $\pi$ (as suboptimal thresholds $\hat\theta$ deviate from optimal thresholds).
\end{assumption}
\begin{assumption}[Treatment effect regime.]
	$$\textstyle (2 \mathbb P(Y= 1\mid 0) -1)  \label{asn-structuralanalysis-teregime}
	+ 
	\E[(2\tau(X) -1 ) \mathbb{I}[{ \outcomeratio(X) > \hat \theta_{t+1}(s)  }]] \leq 0 $$
	
\end{assumption}

\paragraph{\Cref{thm-example-bias-amplification} (full statement for general case, $t<T-2$.).} 

Let $\tau(x)=\eta(1\mid1,x)-\eta(1\mid0,x).$ 	
For $t=T-2$, assume \cref{asn-structuralanalysis-rewardorder,asn-structuralanalysis-rewardorder}. Then, for $s>2$, $$
\textstyle	\E[ -\tau(X)
\mathbb{I} [{\hat{\theta}^*_{T-1}\leq \outcomeratio^*(X) \leq \theta^*_{T-1}}]]\geq0 \implies		{\textstyle \hat\theta_{T-2}^*(\Delta V^{\hpi{T-1:T}}_{T-1})< \hat\theta_{T-2}^*(\Delta V^{\spi{T-1:T}}_{T-1}) }.$$

For $t< T-2$ and $s>2$, assuming \cref{asn-structuralanalysis-concavity,asn-structuralanalysis-rewardorder,asn-structuralanalysis-teregime,asn-structuralanalysis-unimodality}, if 
\begin{align*}
	&(2 V_{t+1}^{\pi^*_{t+1:T}}(s-1)-(V_{t+1}^{\pi^*_{t+1:T}}(s)+V_{t+1}^{\pi^*_{t+1:T}}(s-2))) 
	\E[ -\tau(X)
	\mathbb{I} [{\hat{\theta}^*_{T-1}\leq \outcomeratio^*(X) \leq \theta^*_{T-1}}]]
	\\
	& \qquad 
	+\E_{\pi^*_t(s) -\hat\pi^*_t(s)} [ R(y) + V^{\spi{t+1:T}}_{t+1}(s')\mid s-1 ]- 	\E_{\pi^*_t(s-1) -\hat\pi^*_t(s-1)} [ R(y) + V^{\spi{t+1:T}}_{t+1}(s')\mid s-1 ] \leq 0, 
\end{align*} 
then $		\textstyle \hat\theta_{t}^*(\Delta V^{\hpi{t+1:T}}_{T+1})< \hat\theta_{t}^*(\Delta V^{\spi{t+1:T}}_{t+1})$.

These conditions are problem dependent, depending on the true response models. \Cref{asn-structuralanalysis-concavity} is satisfied for the specific dynamic pricing example, and \Cref{asn-structuralanalysis-unimodality} is a common assumption that is satisfied if conditional responses satisfy log-concave noise distributions. \Cref{asn-structuralanalysis-teregime} is an assumption about the treatment effect and outcome regime, which would be satisfied in a ``sparse reward, not very large treatment effect" regime. 
We collect some lemmas used for the proof. In this section, we focus on the marginal MDP and for clarity in the notation, omit $\tilde V$ and instead write $V_t(S)$ for the value function in the marginal MDP. 
\begin{lemma}[Discrete concavity of the value function]\label{lemma-discreteconcavity}
	
	Suppose $R\geq 0$.	For any $t\in[\horizon+1 ]$ and $\pi$, $V^\pi_t(s)$ is (discrete) concave in $s$. 
	$\\$
\end{lemma}

\begin{lemma}[Single-step deviation and difference in value function differences]\label{lemma-biasanalysis-singlestep} 
	For $t < T-1,$ 
	\begin{align*}
		&	
		\Delta V^{\hat \pi^*_{t},\pi^*_{t+1:T}}_{t}(s)-	\Delta V^{\pi^*_{t},\pi^*_{t+1:T}}_{t}(s)
		\\
		&=	(2 V_{t+1}^{\pi^*_{t+1:T}}(s-1)-(V_{t+1}^{\pi^*_{t+1:T}}(s)+V_{t+1}^{\pi^*_{t+1:T}}(s-2))) 
		\E[ -\tau(X)
		\mathbb{I} [{\hat{\theta}^*_{T-1}\leq \outcomeratio^*(X) \leq \theta^*_{T-1}}]]
		\\
		& \qquad 
		+\E_{\pi^*_t(s) -\hat\pi^*_t(s)} [ R(y) + V^{\spi{t+1:T}}_{t+1}(s')\mid s-1 ]- 	\E_{\pi^*_t(s-1) -\hat\pi^*_t(s-1)} [ R(y) + V^{\spi{t+1:T}}_{t+1}(s')\mid s-1 ]
	\end{align*}
	For $t=T-1$:
	\begin{align*}
		&	
		\Delta V^{\hat \pi^*_{T-1},\pi^*_T}_{T-1}(s)-	\Delta V^{\pi^*_{T-1},\pi^*_T}_{T-1}(s)
		\\
		&=	(2 V_T^{\pi^*_T}(s-1)-(V_T^{\pi^*_T}(s)+V_T^{\pi^*_T}(s-2))) \int (\etaax{0} - \etaax{1})
		\left( \indic{\outcomeratio^* > \theta^*_{T-1}} - \indic{\outcomeratio^* > \hat\theta^*_{T-1}}
		\right)
		dFx
	\end{align*}
\end{lemma}
\begin{lemma}[Value function decomposition]\label{eqn-biasanalysis-sminush}
	Let 	$V^{\hat\pi^*_{t:T}}_t(s)= V^{\pi^*_{t:T}}_t(s) + \delta^V_{t}(s)$. 
	
	\begin{align}&V_{t}^{\spi{t:T}}(s) - V_{t}^{\hpi{t:T}}(s)\\
		&=V_{t}^{\pi^*_{t},\pi^*_T}(s) - V_{t}^{\hat\pi^*_{t},\pi^*_T}(s)
		-\sum_{s',y\mid s} \delta_{t+1}^V(s') \int \left(\etaax{1} \indic{\outcomeratio > \hat\theta^*_{t}} + \etaax{0}(1-\indic{\outcomeratio > \hat\theta^*_{t}} )
		\right) dFx
	\end{align}
\end{lemma}

\begin{proof}[Proof of \Cref{thm-example-bias-amplification}]
	
	First, observe that 
	\begin{equation}\hat\theta^*_{t}(\Delta V^{\hpi{t+1:T}}_{t+1}) \leq \hat\theta^*_{t}(\Delta V^{\pi^*_{t+1:T} }_{t+1}) \iff \Delta V^{\hpi{t+1:T}}_{t+1} \geq \Delta V^{\spi{t+1:T}}_{t+1}
	\end{equation} 
	To see this, consider for simplicity  $ \frac{a+x}{b+x} < \frac{a+y}{b+y}$, where $a=R(0),b=R(1), x=\Delta V^{\hat\pi},y=\Delta V^{\pi^*}$. Note that $a < b$, e.g. $R(0) < R(1)$ and $x> y$ and $\frac{a+x}{b+x} < \frac{a+y}{b+y} \iff (a-b)(y-x) < 0,$
	i.e. iff $\Delta V^{\hat\pi_{t+1:T}}_{t+1}>\Delta V^{\pi_{t+1:T}}_{t+1}$.

	Therefore we will show $ \Delta V^{\hpi{t:T}}_{t}\geq \Delta V^{\spi{t:T}}_{t}, \forall t, s>1$. 
	
	We will also establish that for a given timestep, the bias in thresholds is decreasing in the system state (as are the value function differences). To summarize, we consider the following inductive hypotheses: 
	\begin{align}
		&\Delta V^{\hpi{t+1:T}}_{t+1}\geq \Delta V^{\spi{t+1:T}}_{t+1} \label{eqn-induction-1}\\
		&	\theta_{t+1}(s) - \hat\theta_{t+1}(s) \leq  \theta_{t+1}(s-1) - \hat \theta_{t+1}(s-1)
		\label{eqn-induction-2}\\
		&
		\Delta V^{\spi{t+1:T}}(s) - 		\Delta V^{\hpi{t+1:T}}(s)
		\leq 		\Delta V^{\spi{t+1:T}}(s-1) - 		\Delta V^{\hpi{t+1:T}}(s-1)\label{eqn-induction-3}
	\end{align}

	We prove the inductive step by assuming the above inductive hypotheses are true for $t'=t+1$ and verifying that this implies they hold for $t'=t$. The main analysis is in verifying $\Delta V^{\hpi{t+1:T}}_{t+1}\geq \Delta V^{\spi{t+1:T}}_{t+1}$, \cref{eqn-induction-1}, which requires the other induction hypotheses. We first show the inductive step holds for \cref{eqn-induction-2,eqn-induction-3} under the induction hypotheses. Note that in the special case of $t=T-2$, only \cref{eqn-induction-1} is needed. 
	
	\underline{Inductive step for \cref{eqn-induction-2}.
	}
	To lighten the notation for the following comparisons, we denote $\Delta \widehat{V}^*(s) = \Delta {V}^{\hpi{t:T}}(s), \Delta V^*(s) = \Delta{V}^{\spi{t:T}}(s)$. 
	\begin{align*}&  \theta_t(s)-\hat\theta_t(s) \leq \theta_t(s-1) - \hat\theta(s-1) \\
		&\iff
		\frac{
			(R(0)-R(1)) ( 
			\Delta V^*(s) - \Delta \widehat{V}^*(s) 
			) }{
			(R(1)+ \Delta V^*(s) )(R(1)+\Delta \widehat{V}^*(s) )} > 
		\frac{
			(R(0)-R(1)) ( 
			\Delta V^*(s-1) - \Delta \widehat{V}^*(s-1) 
			) }{
			(R(1)+ \Delta V^*(s-1) )(R(1)+\Delta \widehat{V}^*(s-1) )}\\
		&
		\iff \frac{
			\Delta V^*(s) - \Delta \widehat{V}^*(s) }{
			\Delta V^*(s-1) - \Delta \widehat{V}^*(s-1) 
		} < 
		\frac{ 	(R(1)+ \Delta V^*(s) )(R(1)+\Delta \widehat{V}^*(s) )}{
			(R(1)+ \Delta V^*(s-1) )(R(1)+\Delta \widehat{V}^*(s-1) )}
	\end{align*} 
	where from the second to last line, we use the fact that $R(0)-R(1)<0$ by assumption (without loss of generality) of \Cref{thm-example-bias-amplification} on the ordering of the rewards. 
	
	The LHS of the last inequality above is less than 1 by the induction hypothesis on value function differences in states, $	\Delta V^*(s) - \Delta \widehat{V}^*(s) \leq
	\Delta V^*(s-1) - \Delta \widehat{V}^*(s-1)$. 
	
	The RHS is greater than $1$ since $$(R(1)+ \Delta V^*(s) )(R(1)+\Delta \widehat{V}^*(s) )
	\geq 
	(R(1)+ \Delta V^*(s-1) )(R(1)+\Delta \widehat{V}^*(s-1) ),$$ because \Cref{lemma-discreteconcavity} implies $\Delta V(s)$ is increasing as $s$ increases, so $\Delta V^*(s)\geq\Delta V^*(s-1)$ and $\Delta \widehat{V}^*(s)\geq\Delta \widehat{V}^*(s-1)$.

	\underline{Inductive step for \cref{eqn-induction-3}.
	}

	\Cref{eqn-induction-3} is equivalent to showing 
	$\Delta V^{*}(s) - 		\Delta {V}^{*}(s-1)	
	\leq 	\Delta \widehat{V}^{*}(s) -	\Delta \widehat{V}^{*}(s-1)$. First we decompose the difference into the single-stage reward difference term and the policy-induced transitions to next value functions. Having verified \cref{eqn-induction-2} for time $t$, in combination with \cref{asn-structuralanalysis-unimodality} which implies that greater differences in biased vs. optimal thresholds lead to greater reward suboptimality, implies the single-stage term satisfies the inequality. Verifying the inequality for the difference in value functions term holds by an argument similar to used in showing \cref{eqn-induction-2}, that the region of integration (even after accounting for differences in thresholds) remains one of positive measure, while the integrand is negative (satisfies the inequality) by the induction hypothesis for \cref{eqn-induction-3}, for next-time-step value function differences over states. 
	
	\underline{Inductive step for $ \Delta V^{\hpi{t+1:T}}_{t+1}\geq \Delta V^{\spi{t+1:T}}_{t+1}$}:

	We first establish the base case by studying some properties of $V_T$ and its differences which simplify the analysis. 
	\begin{equation}\label{eqn-biasanalysis-boundary-condition} V^{\pi^*_T}_T(s')-V^{\hat\pi^*_T}_T(s')=V^{\pi^*_T}_T(s) - V^{\hat\pi^*_T}_T(s), \forall s,s' \geq 1
	\end{equation}  
	Therefore $\Delta V^{\pi^*_T}_T(s) - \Delta V^{\hat\pi^*_T}_T(s) 
	= 0$, for $s>1$. 
	
	Note that $\theta^*_T$ is independent of state. Under \Cref{asn-lin-mdp-pot-out}, when $s > 2$, $ V^{\pi^*_T}_T(s-1)-V^{\hat\pi^*_T}_T(s-1)=V^{\pi^*_T}_T(s) - V^{\hat\pi^*_T}_T(s)$, since $V_{T+1}(s)=0$, since the only non-zero terms are invariant in states. The above follows by rearranging. When $s>1$, the statement is also true because next-stage value is $0$ (independent of downstream estimation error), and it is true by definition for $s=0$. 
	
	To establish the inductive step, we decompose $\Delta V^{\pi^*_{t+1:T}}_{t+1} - \Delta V^{\hat\pi^*_{t+1:T}}_{t+1}$.
	By definition,
	\begin{equation}\label{eqn-structuralanalysis-decomposition}
		\Delta V^{\pi^*_{t+1:T}}_{t+1} (s)- \Delta V^{\hat\pi^*_{t+1:T}}_{t+1} (s)= 
		(V_{t+1}^{\spi{t+1:T}}(s-1)-V_{t+1}^{\hpi{t+1:T}}(s-1)) - 	(V_{t+1}^{\spi{t+1:T}}(s)-V_{t+1}^{\hpi{t+1:T}}(s)) 
	\end{equation}
	By \cref{eqn-biasanalysis-sminush},
	\begin{align*}
		&		\Delta V^{\pi^*_{t+1:T}}_{t+1} (s)- \Delta V^{\hat\pi^*_{t+1:T}}_{t+1} (s) \\
		&= 
		\Delta V^{\pi^*_{t+1},\pi^*_T}(s)-\Delta V^{\hat \pi^*_{t+1},\pi^*_T}(s)\\
		&		+
		\sum_{s',y\mid s} \delta_{t+1}^V(s') \int \left(\etaax{1} \indic{\outcomeratio > \hat\theta^*_{t+1}} + \etaax{0}(1-\indic{\outcomeratio > \hat\theta^*_{t+1}} )
		\right) dFx \\
		&		- 	\sum_{s',y\mid s-1} \delta_{t+1}^V(s') \int \left(\etaax{1} \indic{\outcomeratio > \hat\theta^*_{t+1}} + \etaax{0}(1-\indic{\outcomeratio > \hat\theta^*_{t+1}} )
		\right) dFx 
	\end{align*}
	Let $C(y)=\int \left(\etaax{1} \indic{\outcomeratio > \hat\theta^*_{t+1}} + \etaax{0}(1-\indic{\outcomeratio > \hat\theta^*_{t+1}} )
	\right) dFx $. 
	
	Then: 
	\begin{align}&	\Delta V^{\pi^*_{t+1:T}}_{t+1} (s)- \Delta V^{\hat\pi^*_{t+1:T}}_{t+1} (s)
		\\&= 
		\underbrace{	\Delta V^{\pi^*_{t+1},\pi^*_T}(s)-\Delta V^{\hat \pi^*_{t+1},\pi^*_T}(s) }_{\1}+
		\underbrace{		 \delta^V_{T}(s-1) (C(1)-C(0))  + \delta_{T}^V(s) C(0) - \delta_{T}^V (s-2) C(1)}_{{\2}} \label{eqn-structuralanalysis-lemma3decomposition}
	\end{align} 
	We first analyze $\1$, the value function difference under a single-timestep policy difference. By \Cref{lemma-biasanalysis-singlestep},
	\begin{align*}
		&	
		\Delta V^{\hat \pi^*_{t},\pi^*_{t+1:T}}_{t}(s)-	\Delta V^{\pi^*_{t},\pi^*_{t+1:T}}_{t}(s)
		\\
		&=	(2 V_{t+1}^{\pi^*_{t+1:T}}(s-1)-(V_{t+1}^{\pi^*_{t+1:T}}(s)+V_{t+1}^{\pi^*_{t+1:T}}(s-2))) 
		\E[ -\tau(X)
		\mathbb{I} [{\hat{\theta}^*_{t}(s)\leq \outcomeratio^*(X) \leq \theta^*_{t}(s)}]]
		\\
		& \qquad 
		+\E_{\pi^*_t(s) -\hat\pi^*_t(s)} [ R(y) + V^{\spi{t+1:T}}_{t+1}(s')\mid s-1 ]- 	\E_{\pi^*_t(s-1) -\hat\pi^*_t(s-1)} [ R(y) + V^{\spi{t+1:T}}_{t+1}(s')\mid s-1 ]
	\end{align*}

	We establish that the last term is equivalent to integrating over an interval, such that the last term is generally positive since the value functions are nonnegative. Therefore, in the general case, negativity of the first term from the sufficient condition when $t=T-2$ isn't completely sufficient. However, the same inequality may hold given that the state-wise differences in biased threshold suboptimalities is not too large. 
	
	From the inductive hypothesis that $\hat\theta_{t+1}(s) < \theta_{t+1}(s), \forall s>1$, and \Cref{lemma-discreteconcavity} which implies that $\Delta V(s)$ is increasing in $s$ (by properties of discrete derivatives of discrete concave functions), we deduce: 
	\begin{align}
		&\hat\theta_{t+1}(s) \leq \theta_{t+1}(s) \leq \theta_{t+1}(s-1) , \qquad \hat\theta_{t+1}(s) \leq \hat\theta_{t+1}(s-1) \leq \theta_{t+1}(s-1) \label{eqn-structural-analysis-inequalities} 
	\end{align}
	By the properties in \cref{eqn-structural-analysis-inequalities} and the induction hypothesis \cref{eqn-induction-2}, integrating against an increasing function in $\outcomeratio$ is nonnegative: $$
	\E\left[ g(\outcomeratio)
	{\indic{\hat\theta^*_{t+1} (s)< \outcomeratio^* < \theta^*_{t+1}(s)} 			
		-	\indic{\hat\theta^*_{t+1} (s-1)< \outcomeratio^* <	\hat\theta^*_{t+1}  (s)}}
	\right]\geq 0
	.$$

	We next analyze $\2$. 
	Note the inductive hypothesis implies $\delta_{t+1}^V(s-1) \leq \delta_{t+1}^V(s)$, e.g. $\delta_{t+1}^V(s)$ is decreasing in $s$, by rewriting \cref{eqn-structuralanalysis-decomposition} with definition of $\delta^V$. 
	We decompose $C(1)-C(0)$ as: 
	$$ C(1) - C(0) = 
	(2 P(Y(0) = 1) -1) 
	+ 
	\int (2(\eta(1\mid1,x)-\eta(1\mid0,x)) -1 ) \indic{ \outcomeratio > \hat \theta_{t+1}  }dFx
	$$
	Therefore, under \Cref{asn-structuralanalysis-teregime},
	\begin{align*}\2&  = (\delta_T^V(s-1) - \delta_T^V(s)) (C(0)-C(1)) + (\delta_T^V(s-2)-\delta_T^V(s)) C(1) \\
		&\leq 0
	\end{align*}

	\paragraph{Simplification when $t=T-1$.}
	When $t=T-1$, the sufficient condition is a direct consequence of \cref{eqn-structuralanalysis-lemma3decomposition}.
	Since
	$\delta_T^V(s) = \delta_{T}^V (s'), \forall s,s'\in \Ss$, $\2=0$. 
	
	Applying \Cref{lemma-biasanalysis-singlestep},
	\begin{align*}	
		&\Delta V^{\pi^*_{T-1:T}}_{T-1} (s)- \Delta V^{\hat\pi^*_{T-1:T}}_{T-1} (s)\\&
		= 	((V_T^{\pi^*_T}(s)+V_T^{\pi^*_T}(s-2))-2 V_T^{\pi^*_T}(s-1)) 	\E[ -\tau(X)
		\mathbb{I} [{\hat{\theta}^*_{t}(s)\leq \outcomeratio^*(X) \leq \theta^*_{t}(s)}]]
	\end{align*} 
	
	The first multiplicative term is negative by \Cref{lemma-discreteconcavity}, concavity of $V^{\pi^*_T}(s)$ in $s$. 
\end{proof}

\paragraph{Proofs of auxiliary lemmas}

\begin{proof}[Proof of \Cref{lemma-discreteconcavity}]
	This is a structural result of the dynamic pricing problem. We include the proof for completeness but the argument is not novel: we simply verify the adaptation of Theorem 1.18 \cite{gallego2019revenue} holds for the contextual setting in this paper. 
	
	Note that the difference from that formulation is that randomness is modeled in the transition probabilities, not the arrival rates of consumers, and we express $\Delta V(s) = V(s-1)-V(s)$ as the negative finite difference. 
	
	The proof shows concavity of $V_t(s)$ in $s$ by showing that $\Delta V(s)$ is increasing in $s$; hence finite differences (discrete derivatives) are decreasing so that the value function $V$ is concave. The argument follows by forward induction on the state space and a sample path argument. 
	
	The base case holds by definition of $\Delta V(0) = -\infty$; clearly $\Delta V_t(1) > \Delta V_t(0)$ for any $t$. The induction hypothesis posits that for some $s+1$, following the optimal policies, $\Delta V_t^{\pi^*}(s')$ is increasing in $s'$ for $s' \leq s$ for all $t$. We want to show $\Delta V_t^{\pi^*}(s+1)\geq \Delta V^{\pi^*}_t(s)$, or equivalently 
	$$ V_t^{\pi^*}(s+1)+ V^{\pi^*}(t,s-1) \leq V^{\pi^*}(s) + V^{\pi^*}(s) $$
	We verify the sample path argument of \cite{gallego2019revenue} holds in this setting with actions taken in the marginal MDP formulation. We will show:
	$$ V_t^{\pi^*(s+1)}(s+1)+ V^{\pi^*(s-1)}_t(s-1) \leq V^{\pi^*(s+1)}(s) + V^{\pi^*(s-1)}(s)
	$$
	Clearly by suboptimality of the policies optimal at states $s+1, s-1$ for state $s$, $ V^{\pi^*(s+1)}(s) + V^{\pi^*(s-1)}(s)\leq
	V^{\pi^*(s)}(s) + V^{\pi^*(s)}(s)$ so showing the above inequality is sufficient to verify the inductive step.
	
	The sample path argument tracks the usage of the \textit{suboptimal} policies of the left-hand-side original-state $s+1$ and $s-1$ systems, $\pi^*(s+1), \pi^*(s-1)$ for the right-hand-side state-$s$ system, until one of the following cases: $t = T$ (time runs out); at some $t'\geq t$ the difference in inventories of the state-$s+1$ and state $s-1$ systems drops to 1, or the state of the original state $s-1$ system drops to 0. Then the optimal policies for the system state are followed thereafter. 
	
	\underline{Case 1}: Use $\pi^*(s+1), \pi^*(s-1)$ for the two state $s$ systems, respectively, until the end of selling horizon. 
	
	The realized revenues by following the same randomness sample path and same action policies are identical by following the same policies. 
	
	\underline{Case 2}: At some $t'\geq t$ the difference in inventories of the state $s+1$ and state $s-1$ systems drops to 1.
	
	Up to this stopping time, the realized revenues are identical by the stopping path argument. Because the transition realizations are identical, then the right-hand-side systems have the same state space as the left-hand-side systems, since under $\pi^*(s+1)$, $s-s'$ items sold while under $\pi^*(s-1)$, $s-s'-1$ items had sold. At some $t'\geq t$, at some state $s' \leq s$, following optimal policies thereafter, the LHS value functions are given by 
	$$V_{t'}^{\pi^*(s'+1)_{t'}, \pi_{t+1:T}^*}(s'+1) + V_{t'}^{\pi^*(s'+1)_{t'}, \pi_{t+1:T}^*}(s') \geq 
	V^{\pi^*(s')_{t'}, \pi_{t+1:T}^*}(s')+V_{t'}^{\pi^*(s'+1)_{t'}, \pi_{t+1:T}^*}(s'+1)$$
	so that the remaining optimal expected revenues are identical. 
	
	\underline{Case 3}: For some $t' < \horizon$, original state $s-1$ system stocks out, e.g. has $s_{t'}=0$. 
	
	At this point, following policy $\pi^*_t(s+1)$ yields a state $s'$ such that $1 < s' \leq x+1$ (the first inequality holds because otherwise we would be in case 2), so that the states at this timepoint are $(t', s'), (t',0)$. By the sample path argument, identical amounts of goods have sold so the states of the right-hand-side systems are $(t', s'-1), (t', 1)$. By the inductive hypothesis, $\Delta V_{t'}^{\pi_{t'}^*(s')}( s') \geq \Delta V_{t'}^{\pi_{t'}^*(1)}( 1)$ for all $s' \leq s+1$ and all $t' \leq t$. We verify the downstream revenues are at least as high for the right hand systems: 
	
	$$ V_{t'}^{\pi^*}(s') + V_{t'}(0) \leq V_{t'}(s'-1)+V_{t'}(1) $$
	The inductive hypothesis verifies that $\Delta V(s') \geq \Delta V(1)$, which verifies the above. 
	
\end{proof}

\begin{proof}[Proof of \Cref{lemma-biasanalysis-singlestep}]
	Since we restrict attention to single-timestep deviations (following the optimal policy after), we can write $\hat\pi^*(\Delta V^{\spi{t+1:T}}_{t+1}(s))$ in terms of a single threshold based on the optimal value function difference, $\theta^*=\theta^*(\Delta V^{\spi{t+1:T}}_{t+1}(s))$. 
	\begin{align*}&V^{\pi^*_t,\pi^*_{t+1:T}}_t(s) - V^{\hat\pi^*_t,\pi^*_{t+1:T}}_t(s) 
		= \E_{\pi_t^*(s) - \pi_t^*(s)}[R(y) + V^{\pi_{t+1:T}^*}(s')\mid s]
		\\
		&= \sum_{s',y\mid s} \int \mu(y\mid 1,x) (R(y)+V_{t+1}^{\pi^*_{t+1:T}}(s')) \left( \indic{\outcomeratio^* > \theta^*_t(s)} - 
		\indic{\outcomeratio^* > \hat\theta^*_t(s)} \right) dFx \\
		& +  \int \mu(y\mid 0,x) (R(y)+V_{t+1}^{\pi^*_{t+1:T}}(s')) 
		\left( \indic{\outcomeratio^* < \theta^*_t(s)} - \indic{\outcomeratio^* < \hat{\theta}^*_t(s) } \right)
		dFx\\
		& = \sum_{s',y\mid s} \int 
		\left( \mu(y\mid 1,x) (R(y)+V_{t+1}^{\pi^*_{t+1:T}}(s')) -
		\mu(y\mid 0,x)(R(y)+V_{t+1}^{\pi^*_{t+1:T}}(s')) \right)\\
		&  \qquad\qquad	\left( \indic{\outcomeratio^* > \theta^*_t(s)} - \indic{\outcomeratio^* > \hat\theta^*_t(s)}\right)
		dFx 
	\end{align*}
	The claim follows from the above simplification and by expanding the definition, 
	$$		\Delta V^{\hat \pi^*_{T-1},\pi^*_T}_{T-1}(s)-	\Delta V^{\pi^*_{T-1},\pi^*_T}_{T-1}(s)
	=(V^{\pi^*_{t},\pi^*_{t+1:T}}_{t}(s) - V^{\hat\pi^*_{t},\pi^*_{t+1:T}}_{t}(s) )- (V^{\pi^*_{t},\pi^*_{t+1:T}}_{t}(s-1)-V^{\hat\pi^*_{t},\pi^*_{t+1:T}}_{t}(s-1)),$$
	and algebraic manipulation of the resulting expression. 
	The statement for $t=T-1$ follows since $\theta_{T-1}$ is the same for all states, so that the reward differences also cancel out when the differences of $\Delta V$ are considered. 
	
	For $t < T-1$, relative to the simplification of $T-1$ we obtain an additional term that arises from the differences in $\theta_{t+1}(s)-\hat\theta_{t+1}(s)$ for different states $s$: 
	$$ \E_{\pi_t^*(s) - \pi_t^*(s)}[R(y) + V^{\pi_{t+1:T}^*}(s')\mid s] - 
	\E_{\pi_t^*(s) - \pi_t^*(s)}[R(y) + V^{\pi_{t+1:T}^*}(s')\mid s-1] $$
\end{proof}

\begin{proof}[Proof of \cref{eqn-biasanalysis-sminush}]

	\begin{align*}
		&		V_{t}^{\pi^*_{t:T}}(s) - V_{t}^{\hat\pi^*_{t:T}}(s) \\
		&= \sum_{s',y\mid s} \int (\etaax{1}(R+V^{\pi^*_{t+1:T}}_{t+1}(s')) \indic{\outcomeratio > \theta^*_{t}} + \etaax{0} (R + V^{\pi^*_{t+1:T}}_{t+1}(s')) )  (1-\indic{\outcomeratio > \theta^*_{t}})  dFx \\
		&		- \sum_{s',y\mid s} \int (\etaax{1}(R+V^{\pi^*_{t+1:T}}_{t+1}(s')+ \delta^V_{t+1}(s'))\indic{\outcomeratio> \hat\theta^*_{t}}  
		+ \etaax{0} (R + V^{\pi^*_{t+1:T}}_{t+1}(s')+\delta^V_{t+1}(s')) ) 
		\\&\qquad \qquad 
		\cdot(1-\indic{\outcomeratio > \hat\theta^*_{t}} ) dFx
	\end{align*}
	and collect terms corresponding to $V_{t}^{\pi^*_{t:T}}(s) - V_{t}^{\hat\pi^*_{t:T}}(s)$.
\end{proof}

\section{Discussion}\label{apx-sec-discussion} 

\subsection{Further Related Work}\label{apx-sec-relatedwork}

\textbf{Online contextual decision-making with constraints.} 
There is an extensive literature on either contextual or stateful problems in operations research, including online learning. Typically contexts are discrete, known types. We highlight work that studies online learning in constrained systems, such as (episodic) inventory/revenue management, \citep{huh2011adaptive,besbes2012blind,agrawal2019learning}, or contextual decisions such as covariate-based dynamic pricing \cite{cohen2016feature,javanmard2016dynamic,qiang2016dynamic,shah2019semi,ban2020personalized,chen2021statistical}. These approaches are typically model-based: they require uncontextual demand distributions (known, or learned online) or impose parametric restrictions. Contextual bandits with knapsack (CBwK) does consider both contexts and statefulness.
\cite{badanidiyuru2018bandits}.\footnote{We discuss CBwK for a full discussion of related work. But while our framework can readily handle unknown or multiple behavior policies, we do not consider data directly collected from a bandit algorithm (i.e. outcome-adapted data subject to adaptive sequential learning bias).} The closest work is \cite{agrawal2016efficient},
which uses
single-timestep offline policy optimization 
but considers the Lagrangian relaxation of the resource constraints: regret guarantees are on the Lagrangian and the policy satisfies constraints in expectation rather than with probability 1.

In contrast to the online setting where completely randomized exploration is possible, we are interested in characterizing the setting of learning a dynamic policy from \emph{offline} off-policy data, \textit{without} the ability to set an exploration policy to collect more information. Relative to CBwK and pricing bandits, we consider a general MDP embedding and our sample complexity analysis and algorithm do not require specific structure of the reward beyond \cref{asn-X-exogeneous-context,asn-lin-mdp-pot-out,asn-product-state-policy}.

\textbf{Algorithmic analysis under known distributions.} 
Algorithmic analysis, building on online/approximation algorithms and approximate dynamic programming also requires known demand distributions, hence is complementary
\cite{gallego2019revenue}. Our approach is particularly beneficial in handling high-dimensional context variables $X_t$. Naive extensions of these approaches, for example applying them to an MDP with state aggregation on $X_t$, incurs statistical bias in general due to discretization. 
On the other hand, using \textit{action-history dependent} policies achieves stronger regret guarantees in recent work, e.g., that include resolving (model-predictive control) \cite{bumpensanti2020re}. In contrast, we restrict to state- and time-dependent, but history-independent, policy specifications.

\paragraph{Off-policy policy learning leveraging off-policy evaluation.}
We also compare to backwards-recursive off-policy learning approaches in the dynamic treatment regime literature. In some sense, the DTR/longitudinal causal inference is the opposite of our setting: the difficulty arises from \textit{longitudinal dynamics of the same individual}.  \cite{zhang2013robust} studied an AIPW estimator in the dynamic treatment regime setting but handle policy-dependent nuisances by approximating with a $Q$ function optimized by another method. \cite{zhao2015new} proposes ``backwards outcome-weighted learning" which considers backwards induction on an inverse-propensity weighted estimator that conducts importance sampling in the space of trajectories. Their direct consistency analysis of the backwards induction incurs exponential dependence on horizon. 

\paragraph{Clarification to other settings.} 
\cite{emek2020stateful} introduces ``stateful online learning'', a version of online adversarial learning with state information, but their setting is different. In particular, they focus on MDPs with deterministic transitions and assume bounded-loss simulatability from any state, focusing on the adversarial setting. Our focus on \textit{offline contextual decision-making} with state information is different from the contextual MDP model where contexts index MDP models themselves. 

\subsection{Additional examples of stateful problems}

\begin{example}[Multi-item network revenue management]\label{ex-multinrm}
	Multi-item network revenue management is easily modeled as a modification of \Cref{ex-singlenrm} with additional outcomes (products). Consider a setting with $J$ different products and $K$ many resources, so that $M \in \mathbb{R}^{K\times J}$ is the resource consumption matrix, where $M_{ij}$ describes how much of resource $i$ product $j$ requires. 
	Denote the event $
	\textstyle\mathbb{I}[ s \text{ feas. for }j   ]
	\defeq \underset{i\in [M]}{\prod} \mathbb{I}[M_{ij} < s_i ] $, which describes  the event that the state variable $s$ is feasible to produce product $j$.

	We suppose a joint distribution on $(X,Z)$, e.g. we have exogenous context arrivals and exogenous $Z \mid X$ product types (which may be conditional on the context in the most general case). Therefore at each timestep we sell at most one product at a time. 
\end{example}

The multiple product $Q_t(s, x, j,a)$ function on the expanded state space (including product arrival type) is analogous. In this case, the context-marginalized value notation, $ \tilde{V}^\pi_{t} (S),$ is overloaded: it now marginalizes over the joint distribution of contexts and product types. 

\begin{example}[Pricing and repositioning]
	We adapt a simplified example of setting rental price for vehicles at beginning of each period in a finite (or possibly infinite) planning horizon to a contextual setting \cite{el2020lookahead}. Repositioning is achieved by setting prices to induce directional demand. Discrete state space $\mathcal S $ denotes the number of cars at a station, with $\overline{s}$ the maximum number of cars in vehicle sharing system. Between locations there is a known origin-destination transition probability $\phi_{ij}$. $Y_{ik,t+1}$ is a random variable taking values in $[N]$ that represents the random destination of customer $k$ at station $i$; observed at the beginning of period $t+1$. Uncontextually, $Y_{ik,t+1} = j \text{ w.p. }  \phi_{ij}$. Contextually, we consider $(X_t, O_t, D_t)$ exogeneous covariate and origin-destination request, and the individual demand is a binary outcome in response to price, $Y(p_{it})$. To determine the cost function, let $\ell(i,j)$ be the distance from station $i$ to $j$, and consider a lost sales unit cost $\rho_i, i \in [N]$. 
	The decision vector $\mathbf{p}_t = \{ p_{it} \in [\underline{p}_i , \overline{p}_i], \forall i \in [N] \}$ sets prices for each station. To instantiate the key assumption in this setting, our stateful formulation holds if we believe that the underlying system state $S_t$ is not a confounder because it does not affect whether or not an \textit{individual} demand arrival responds to price. 
	
\begin{align*}
	&V^{*}\left({S}_{t}, X_t, (O_t, D_t)\right)=\\
&	\max _{\mathbf{p}_{t}} 
	\underbrace{
		\E[p_{it} \mathbb{I}[Y=1] \mid X_t, p_{it},O_t, D_t ] \ell(O_t, D_t)
	}_{\text{spatial pricing} }
	-
	\underbrace{
		\rho_{O_t} \mathbb{I}[S_t(O_t) = 0] }_{\text{lost sales penalty}}+
	\gamma 
	\E[ \tildeV^{*}\left({S}_{t+1}\right) \mid p_{it}, X_t, O_t, D_t ] 
\end{align*}

\end{example}
\subsection{Additional comparisons for \Cref{thm-policy-learning-sample-complexity-uniform-convergence}}
\begin{remark}[The constants of \Cref{thm-policy-learning-sample-complexity-uniform-convergence} improve upon uniform concentratability]
	We simply observe that the direct analysis of the above problem-structure-dependent approach improves upon generic application of general algorithms and general problem-independent bounds. A key quantity that appears in batch RL generalization bounds, with different variations, is the \textit{concentratability coefficient}. They originated in \cite{munos2003error} and are used to generally quantify the degree of exploration in the underlying observational data (behavior policy). 
	
	In this setting, {uniform} concentratability coefficients are unnecessarily conservative to describe the degree of required exploration, since the $X_t$ process is exogenous of state dynamics and correspondingly, the behavior policy is not required to strongly explore states. This comparison holds under a favorable observation model, where (as in revenue management) the contextual response arrival/demand is observed even if the system transition is infeasible. 
	We provide a simple example of a stateful MDP with a (uniform, per-timestep) concentratability coefficient that grows exponentially with horizon length. 
	\begin{definition}[Concentratability coefficient]\cite{munos2003error,antos2007fitted,munos2007performance,munos2008finite,le2019batch}.
		
	\end{definition}
	Let $P^{\pi}$ be the operator acting on $f: S\times X \times A \mapsto \mathbb{R}$ s.t. $${\left(P^{\pi} f\right)(s,x, a)=\int_{\Ss\times \Xx} f\left(s^\prime, x^{\prime}, \pi\left(s^\prime, ,x^{\prime}\right)\right) p\left(ds^\prime, d x^{\prime} \mid s,x, a\right) }.$$ Let $d_m\mu$ denote the distribution of states induced by the behavior policy $\mu$ by time $m$. Given data generating distribution $\mu$, under an initial state distribution, for $m \geq 0$ and an arbitrary sequence of stationary policies $\left\{\pi_{m}\right\}_{m \geq 1}$ define the $m$-step concentratability coefficient:$$\beta_{\mu}(m)=\sup _{\pi_{1}, \ldots, \pi_{m}} \left\| \frac{d\left( P^{\pi_{1}} P^{\pi_{2}} \ldots P^{\pi_{m}}\right)}{d \mu} \right\|_{\infty}$$
	In the infinite-horizon setting, it is assumed $\beta_{\mu}=(1-\gamma)^{2} \sum_{m \geq 1} m \gamma^{m-1} \beta_{\mu}(m)<\infty$, e.g. that finiteness arises after discounting. A comparable direct comparison would require an infinite-horizon embedding of a stateful MDP or a finite-horizon analysis of error propagation. We instead study the dependence of $\beta(m)$ in $m$ in a finite horizon version.  
	
	\begin{example}[Uniform concentratability coefficient grows exponentially in horizon for stateful MDPs]
		Consider a simple example with dynamic pricing under capacity constraint $s_0$ with uniform demand.\footnote{Under a restricted amount of covariate-conditional heterogeneity, similar results hold.}The initial state is $s_0 \leq \horizon$. 
		
		To fix intuition, first consider the case of two actions $\Aa = \{ 0,1\}$, where $$
		\pr(Y=1\mid 0)=1-p, \qquad \pr(Y=1\mid 1)=p, \qquad p << \nicefrac 12.$$
		Suppose the behavior policy $\mu$ is such that $\pr(A=0)=1$. Then for the policy $\pr(A=1)=1$, the density ratio at $S_t=s_0,$ is the likelihood ratio of $0$ sales in $\horizon$ timesteps, $$\nicefrac{\nu_T(s_0)}{\mu_T(s_0)}= \nicefrac{(1-p)^T}{p^T}= ( \nicefrac{1}{p} - 1)^T. $$ Therefore $\beta_\mu(T) > \left( \nicefrac{1}{p} - 1\right)^T$, the uniform concentratability coefficient is lower bounded by a constant that is exponential in $\horizon$. 
		
		The previous case provides the main intuition. Now suppose $\pr_\mu(A=0) = 1-\epsilon$ and the evaluation policy satisfies $\pr(A=1)=1-\epsilon$. Note the number of times action $a$ is taken, the random variable $N_A$, is distributed as $N_A(\mu) \sim B(T, \epsilon), N_A(\nu) \sim B(T, 1-\epsilon)$. Consequently the number of selling events is a conditional binomial distribution. $N_Y \mid N_A(\pi) \sim B(N_A(\pi), y(1) )+B(T-N_A(\pi), y(0))$. By properties of the conditional binomial distribution, $$N_Y(\pi)\sim B(T,  \pi(1) y(1)) + B(T, (1-\pi(1))y(0)).$$
		Therefore $$\frac{\nu_T(s_0)}{\mu_T(s_0)} = \frac{\pr(N_Y(\nu) = 0)}{\pr(N_Y(\mu) = 0)} = 
		\frac{(1-\epsilon)^T (1-p)^T + \epsilon^T p^T }{ (1-\epsilon)^Tp^T + \epsilon^T (1-p)^T  }.$$ Choosing $\epsilon < \frac{\tilde \epsilon}{T}$ and considering a Taylor expansion of $(1-\epsilon)^T = 1 - T\epsilon + O(\epsilon^2)$ shows that $$\frac{\nu_T(s_0)}{\mu_T(s_0)} = \frac{ (1-\tilde\epsilon) (1-p)^T + O(\tilde{\epsilon}^2) + O(\tilde{\epsilon}^T) }{ (1-\tilde\epsilon)p^T + O(\tilde{\epsilon}^2)+ O(\tilde{\epsilon}^T) },$$ and hence the leading term is $(1-\tilde\epsilon)(1-\frac{1}{p})^T$. Therefore the uniform concentratability coefficient is exponential in $T$, while single stage-wise overlap is linear in $T$. 
	\end{example}
	
Other refinements of concentratability coefficients have been studied and may obviate this comparison. While \cite{scherrer2014approximate} studies refinements of the uniform concentration coefficient that consider averages over the MDP dynamics, as  \cite{chen2019information} notes, the constant becomes problem-dependent and there are not a priori bounds available for these refined notions. Lastly, it should be noted that other analysis, for example for linear function approximation, studies concentratability in combination with function approximation, e.g. \cite{duan2020minimax} achieves finite bounds for linear function approximation (by virtue of the function approximation).

The goal of the analysis here is to highlight that the necessity of accounting for differences in state distribution is not relevant in our setting, because our algorithms re-use data across states. 

\end{remark}

\subsection{Extension to continuous transition probabilities}\label{apx-sec-cont-transition} 
\begin{remark}%
	Extending to continuous-valued transitions
	further requires estimating the entire \textit{counterfactual distribution} $P(Y(a)=y\mid X=x), \forall y$, a nonstandard task, and allowing for function approximation of policies in continuous $s$. The recent work of \cite{kennedy2021semiparametric}, which derives an efficient and doubly robust estimator, can be used to address the first difficulty; the direct analogue of our approach using that estimator admits doubly-robust off-policy evaluation. It would, however, remain to develop function approximation and a uniform guarantees for the policy learning task. We leave this direction for future work. 
\end{remark}

\subsection{Societal impact statement}\label{apx-sec-socialimpact} 

With regards to \textit{negative societal impacts}, the work is methodological and so its impacts, whether positive or negative, are domain-dependent. However, our key insight was, after all, specializing our focus to systems with exogenous arrivals of \textit{individuals}.  Since this subclass of problems includes resource allocation, it is important to consider fairness concerns about what guarantees we can provide to these {individuals}. In general, system actions may also affect individual utility so that system-wide objectives and reward specifications do not capture individual-level effects. System-optimal policies may also have disparate impacts on different individuals due to differences in contextual responses or even arrival structure. For example, optimizing for the arrival of different customer types with different revenue prospects may have implications for service availability for others. The exact manifestations of these concerns, whether or not they correspond to positive or negative impacts, and possible remedies, must be carefully considered in the particular domain of application.

\section{Experiments}\label{apx-sec-experiments} 

\subsection{Supplementary experiments}

\begin{figure}
	\begin{subfigure}{0.5\textwidth}
	\centering
	\includegraphics[width=\textwidth]{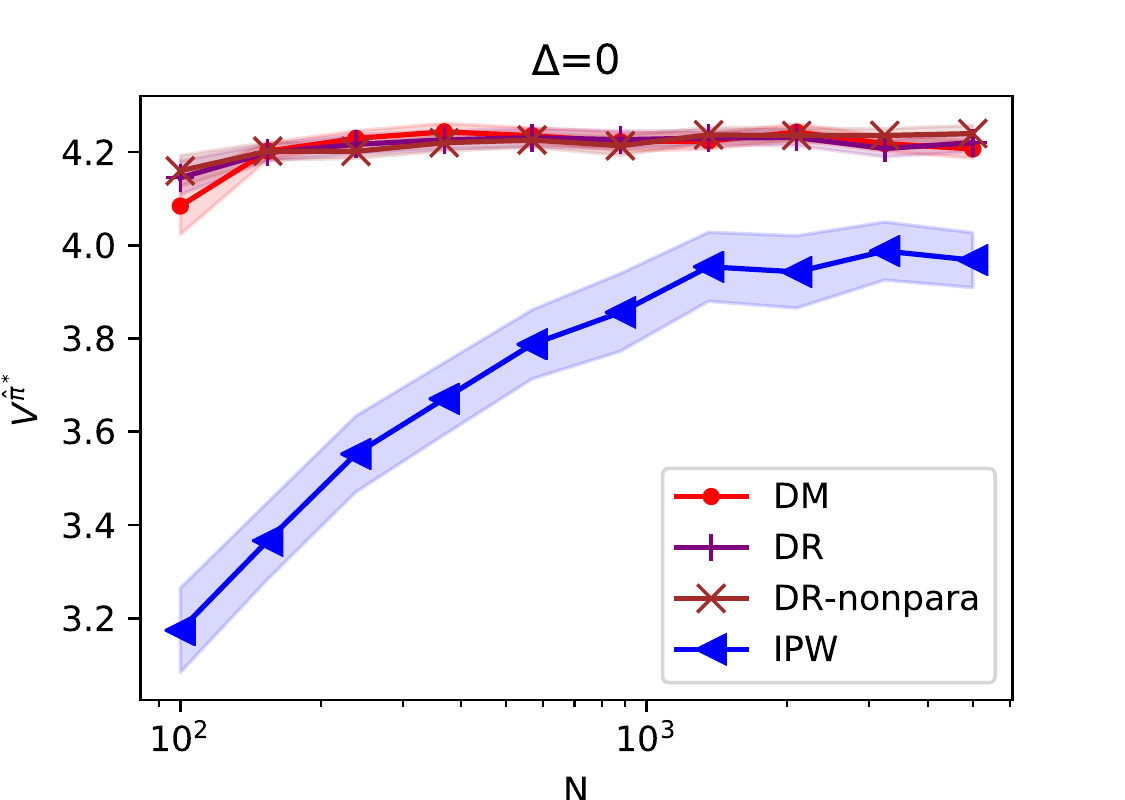}
	\caption{Out-of-sample value $V^{\hat\pi^*}$ (y-axis) of threshold policies optimized via \Cref{alg-backwards-recursion}.}
	\label{fig-polopt-del0}
\end{subfigure}
\caption{Policy evaluation and optimization as more trajectories (x-axis) are collected of $\horizon$-horizon selling in contextual and capacitated dynamic pricing, \cref{ex-singlenrm}; specification of \cref{eqn-dgp}.}
\label{fig-ope-del0}
\end{figure}
\paragraph{IPW in example of \Cref{fig-polopt}}
For completeness we also include the results from the well-specified case, corresponding to $\Delta=0$ in \Cref{fig-polopt}. In the well-specified model case, model based approaches improve handily upon the high-variance $\op{IPW}$ approach.

\paragraph{Additional strong baseline comparisons: MIS and DRL (with marginal state-action density ratios).} 
We also compare to state-of-the-art methods for evaluation. We emphasize that the core contribution of this approach is not in the area of off-policy evaluation itself, but rather policy optimization. Indeed, the proposal for evaluation can be understood as a special case of the backwards-recursive presentation in \cite{jl16}. 

In this setting, we can compare to approaches which do not use the ``stateful" structure directly, as well as adaptations of approaches which mildly adapt to the structure. 
	Let $\rho^{\pi}_{t}(s, x)$ be the distribution of state $S_t=s, X_t=x$ when executing policy $\pi$, starting from initial state $S_0$ drawn from an initial distribution over states. We can correspondingly define a state-action distribution. 
One way to use the ``stateful" structure is to recognize that the finite-horizon \textit{state-action density ratio} $$h_t(s_t,x_t,a_t) = \frac{\rho_{t}^{\pi_e}(s_t,x_t,a_t)}{\rho_{t}^{\pi_b}(s_t,x_t,a_t)}
= 
\frac{\rho_{t}^{\pi_e}(s_t,a_t)}{\rho_{t}^{\pi_b}(s_t,a_t)} \frac{f(x_t)}{f(x_t)}
$$ is \textit{independent of $x$} since $x$ is exogenously generated. Therefore the marginal density ratio only varies in $s_t, a_t$, $h_t(s_t,a_t) = \frac{\rho_{t}^{\pi_e}(s_t,a_t)}{\rho_{t}^{\pi_b}(s_t,a_t)}$. The corresponding marginalized importance sampling estimator of \cite{xie2019towards} is $\E_N[\sum_{t=0}^\horizon h_t(S_t,A_t)R_t ]$. With a nuisance estimate of $Q_\pi$, the efficient estimator of \cite{kallus2019double} is $$\E_N\left[\sum_{t=0}^\horizon h_t(S_t,A_t)(R_t - Q_t^\pi(S_t,X_t,A_t))  - h_{t-1}(S_{t-1},A_{t-1})\E_{a \sim \pi_e}[Q_t(S_t, X_t, a)]  \right].$$
We can compare to strong baseline versions of the above by encoding the structural information by restricting attention to density ratios $h_t(s,a)$ satisfying functional specifications that are independent of $x$. We use $Q$ functions learned by fitted Q-evaluation as nuisance estimates for the method of \cite{kallus2019double}. 

\begin{figure}[t!]
		\begin{subfigure}{0.4\textwidth}
		\centering
		\includegraphics[width=\textwidth]{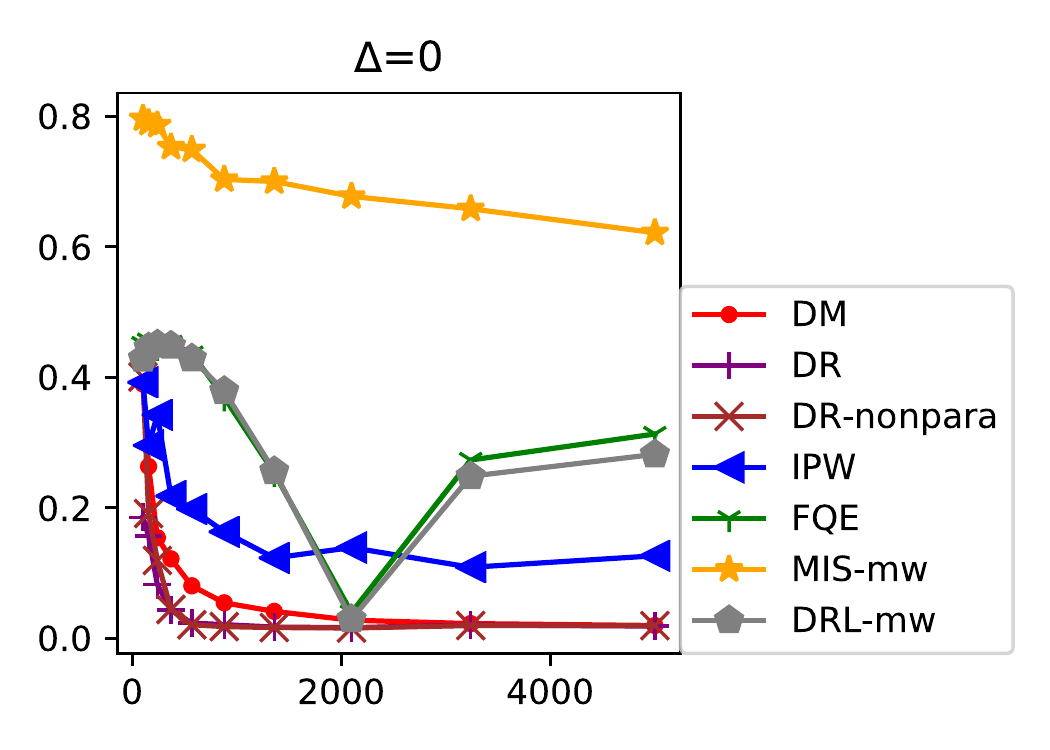}
		\caption{Policy evaluation: $\Delta=0$}
		\label{fig-poleval-higherdim-nu-0}
	\end{subfigure}
	\begin{subfigure}{0.3\textwidth}
		\centering
		\includegraphics[width=\textwidth]{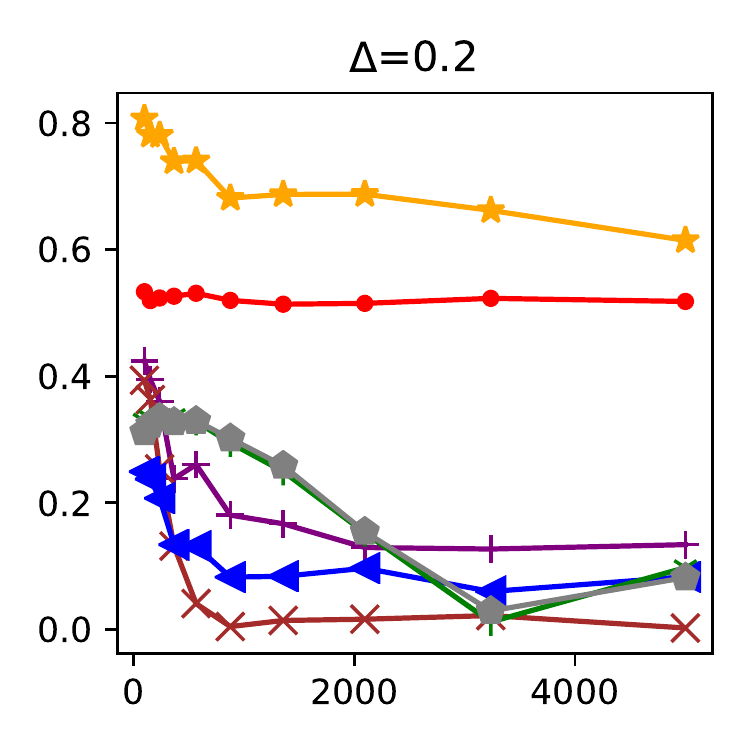}
		\caption{Policy evaluation: $\Delta = 0.2$}
		\label{fig-poleval-higherdim-nu-0.2}
	\end{subfigure}\begin{subfigure}{0.4\textwidth}
		\centering
		\includegraphics[width=\textwidth]{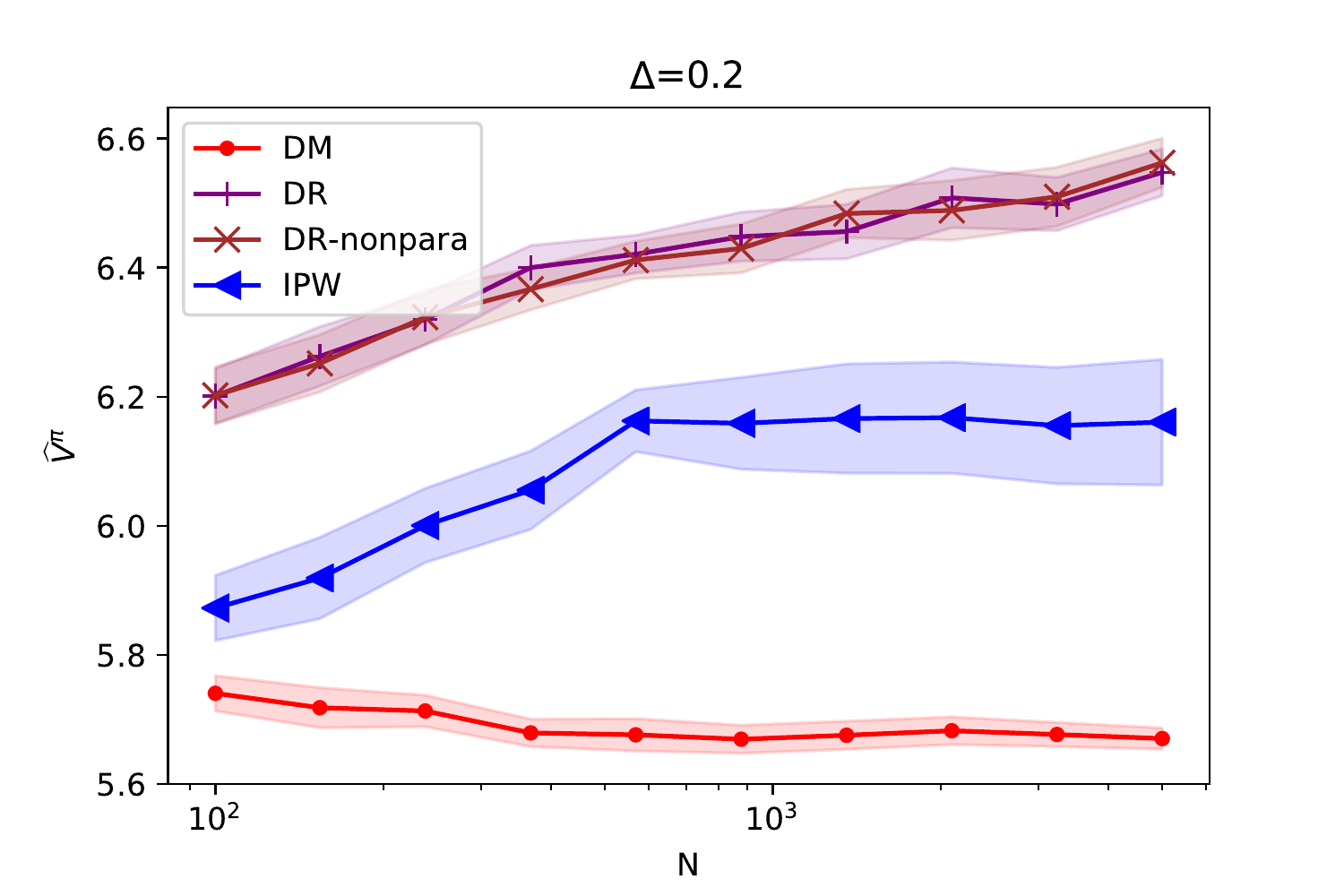}
		\caption{Out-of-sample value $V^{\hat\pi^*}$ (y-axis) of policy optimization: threshold policies optimized via \Cref{alg-backwards-recursion}. Higher is better.}
		\label{fig-polopt-higherdim-sub}
	\end{subfigure}
	\caption{Policy evaluation and optimization as more trajectories (x-axis) are collected of $\horizon$-horizon selling in contextual and capacitated dynamic pricing, \cref{ex-singlenrm}; specification of \cref{eqn-dgp-higherdim}.}
	\label{fig-polopt-higherdim}
\end{figure}

\paragraph{Larger-dimensional policy optimization comparison}
With $p=5$, fix the parameters of the dgp as: 
\begin{equation*}
	\mu = [1 , -0.5, 0.5,  -1, 1], \beta = [2,2,2,-2,-2], \beta_0=-2. 
\end{equation*}
The outcome model and misspecified outcome model are the following: 
\begin{align}
	\textstyle 
	\mu(1\mid a,x) = (1-\Delta) \sigma(\beta^\top x + \beta_0 
	\cdot p(a) )
	+ \Delta\sigma(
	(x_0^2 - x_3 x_4-x_1^2)
	\cdot p(a) ), \;
	e_t(1\mid x)={\sigma(-0.8\beta^\top x )} \label{eqn-dgp-higherdim}
\end{align}
In \Cref{fig-polopt-higherdim} we repeat the comparisons of \Cref{fig-poleval,fig-polopt} and find similar qualitative results and comparisons as to those of \Cref{fig-poleval,fig-polopt}. We repeat over 48 replications. Our approaches that leverage the structure enjoy finite-sample improvements from much simpler nuisance estimation. Double-robustness greatly improves finite-sample performance of unbiased IPW, though well-specification of the nuisance parameter is important. In \Cref{fig-poleval-higherdim-nu-0} we see the benefits of rate-double-robustness even in the well-specified nuisance case: both doubly-robust approaches achieve improve finite sample performance for small $N$, although these benefits become comparable when $N=2000$. We report the median because the fitted-Q-evaluation results in outliers for OPE performance. 

\subsection{Details on computation}\label{apx-sec-compdetails} 
\begin{itemize}
	\item Experiments done on a MacBook Pro with 16gb RAM. 
	\item Experiments develop a custom Gym environment \cite{brockman2016openai} as well as use Numpy, Scipy, Sci-kit Learn \cite{pedregosa2011scikit}. 
\end{itemize}

\end{document}